\documentclass[%
 reprint,
 amsmath,amssymb,
 aps,
twocolumn
]{revtex4-2}

\usepackage{graphicx}
\usepackage{dcolumn}
\usepackage{bm}
\newcommand{\pt}[1]{\left ( #1 \right )}
\newcommand{\pq}[1]{\left [ #1 \right ]}
\newcommand{\pg}[1]{\left \{ #1 \right \}}
\newcommand{\pa}[1]{\left \langle #1 \right \rangle}

\newcommand{\evalat}[2]{\left. #1 \right|_{#2}}

\usepackage{xcolor}
\usepackage{booktabs}
\usepackage{hyperref}
\usepackage{makecell}
\definecolor{myblue}{RGB}{12, 12, 158}
\definecolor{blue(ncs)}{rgb}{0.0, 0.53, 0.74}
\definecolor{crimson}{rgb}{0.86, 0.08, 0.24}
\definecolor{denim}{rgb}{0.08, 0.38, 0.74}
\definecolor{hanblue}{rgb}{0.27, 0.42, 0.81}
\definecolor{cadmiumorange}{rgb}{0.93, 0.53, 0.18}

\usepackage{bm}
\usepackage[normalem]{ulem}
\usepackage{comment}

\begin{document}

\preprint{APS/123-QED}

\title{Unsupervised hierarchical clustering using the learning dynamics of RBMs}

\author{Aurélien Decelle}
 \affiliation{Departamento de Física Teórica, Universidad Complutense de Madrid, 28040 Madrid, Spain.}
 \affiliation{Université Paris-Saclay, CNRS, INRIA Tau team, LISN, 91190, Gif-sur-Yvette, France.}
 
\author{Lorenzo Rosset}%
 \affiliation{Departamento de Física Teórica, Universidad Complutense de Madrid, 28040 Madrid, Spain.}
 \email{lrosset@ucm.es}

\author{Beatriz Seoane}
 \affiliation{Departamento de Física Teórica, Universidad Complutense de Madrid, 28040 Madrid, Spain.}
 \affiliation{Université Paris-Saclay, CNRS, INRIA Tau team, LISN, 91190, Gif-sur-Yvette, France.}

\date{\today}

\begin{abstract}
Datasets in the real world are often complex and to some degree hierarchical, with groups and sub-groups of data sharing common characteristics at different levels of abstraction. Understanding and uncovering the hidden structure of these datasets is an important task that has many practical applications. To address this challenge, we present a new and general method for building relational data trees by exploiting the learning dynamics of the Restricted Boltzmann Machine (RBM). Our method is based on the mean-field approach, derived from the Plefka expansion, and developed in the context of disordered systems. It is designed to be easily interpretable. We tested our method in an artificially created hierarchical dataset and on three different real-world datasets (images of digits, mutations in the human genome, and a homologous family of proteins). The method is able to automatically identify the hierarchical structure of the data. This could be useful in the study of homologous protein sequences, where the relationships between proteins are critical for understanding their function and evolution. 
\end{abstract}

\maketitle


\section{\label{sec:intro}Introduction}

The advent of new technologies is fostering the collection of data in unprecedented quantities. The difficulty is to automatically extract information or discover core principles from this collected data, which is frequently neither labelled nor classified. Biology provides a key illustration of this situation: modern sequencing techniques generate vast amounts of raw data (e.g. genome or protein sequences) far more quickly than researchers can connect it to biological knowledge through experimental procedures or curated annotations. Building the theoretical foundation needed to extract scientifically meaningful insights from massive datasets is more important than ever. In this study, we focus on the problem of unsupervised identification of categories and subcategories in the data without introducing interfering prior knowledge that could bias the results.

Energy-based models in unsupervised learning are particularly well suited to interpretability applications because their operating mechanism is precisely to encode the empirical distribution of the data in the Boltzmann distribution of an energy function (often called {\em Hamiltonian}). If simple enough, such a Hamiltonian serves as an effective model for the data, and can be analyzed for feature extraction. This strategy has been very popular in biology, for instance for protein-structure prediction~\cite{weigt2009identification,morcos2011direct,cocco2018inverse}.
In this sense, Restricted Boltzmann Machines (RBMs) are very well suited for this task because they are powerful enough to encode any complex data set~\cite{le2008representational}, and at the same time simple enough (formally analogous to a disordered spin Hamiltonian in physics) to be described with the tools of statistical physics. In fact, significant progress has been made in the last decade in understanding the learning mechanisms of the RBM, which has provided a theoretical basis for extracting statistical features in these models~\cite{decelle2017spectral,decelle2018thermodynamics,decelle2021equilibrium,tubiana2017emergence,tubiana2019learning}.

Today we know that learning in an RBM is triggered when the eigenvectors of its coupling matrix gradually match the first principal component analysis (PCA) directions of the dataset~\cite{decelle2017spectral,decelle2018thermodynamics,decelle2021equilibrium}. As learning progresses, the RBM finds more complex nonlinear decompositions of the data, which are critical to producing increasingly reliable fake data. The goal is then to use these features to go beyond the aforementioned PCA for dimensionality reduction to identify hidden communities, or features, in the data. In addition, one can exploit the  progressive nature of learning to hierarchize those features: the machine first learns global features to gradually encode more local and fine-grained features that can help us to identify sub-communities. At this point, the simplicity of the RBM architecture provides a clear advantage over more complex neural networks, as the log-likelihood of the models at different stages of the learning process can be easily explored using standard tools of statistical physics of disordered systems, such as the TAP equations to find different probability maxima~\cite{gabrie2015training,tramel2018deterministic,maillard2019high}.
In this work, we implement this idea in real-world datasets, exploiting current knowledge of the learning mechanisms of RBMs to build relational trees with data. We then use these hierarchies to divide the data into families and subfamilies.

Our work is structured as follows: We begin with an introduction to the RBM model and its learning mechanisms. This is followed by a mean-field  description of this model, which is used to find the multiple maxima of the model's probability distribution function that are further used to cluster the data at different stages of the learning process. Later, we explain how we exploit the dynamics of these fixed points to build a tree with a dataset in which categories emerge naturally as learning encodes finer and finer features in the model. Finally, we show the implementation of this algorithm on several real-world datasets.

\section{Definition of the Model}
\begin{figure}[t!]
    \centering
    \includegraphics[width=0.4\textwidth]{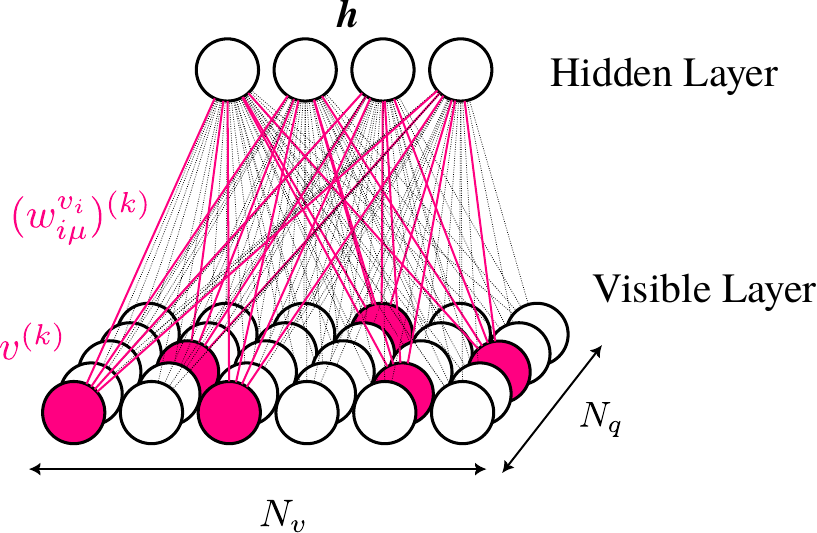}
    \caption{Scheme of the Potts-RBM. The weights matrix is a 3-dimensional tensor $\pmb{w} = \{w_{i \mu}^{q}\}$. When an input data $\pmb{v}^{(k)}$ enters the machine, it selects at each site $i$ the connection of the proper colour $v_i=q$ with the hidden component $\mu$.}
    \label{fig:Potts RBM}
\end{figure}

The RBM is a Markov random field with pairwise interactions defined on a bipartite graph of two non-interacting layers of variables: the visible variables $\pmb{v} = \{v_i\}_{i=1, \dots, N_v}$ represent the data, while the hidden variables $\pmb{h} = \{h_\mu\}_{\mu = 1, \dots, N_h}$ constitute a latent representation of the data that models the effective interactions among the visible variables. The joint probability distribution of visible and hidden variables is given by the Boltzmann distribution
\begin{equation}
    p_{\pmb{\theta}}(\pmb{v}, \pmb{h}) = \frac{1}{Z_{\pmb{\theta}}} e^{- E(\pmb{v}, \pmb{h}; \pmb{\theta})}, \text{  where  } Z_{\pmb{\theta}} = \sum_{\pmb{v}, \pmb{h}} e^{- E(\pmb{v}, \pmb{h}; \pmb{\theta})}.
    \label{eq:Boltzmann distribution}
\end{equation}
In the previous expressions, the normalization factor $Z_{\pmb{\theta}}$ is called the {\em partition function}, $\pmb{\theta}$ refers to the parameters of the model and $E$ is the energy function or Hamiltonian.
In this work, we focus on a setting in which the hidden units are binary variables, $h_\mu \in \{0, 1\}$, while the visible units are categorical variables that can assume values $q \in \{1, \dots, N_q\}$.
We can model this kind of system by introducing a set of \textit{Potts variables} in the visible layer; the associated Hamiltonian is
\begin{equation}\label{eq:hamiltonian}
    E(\pmb{v}, \pmb{h}; \pmb{\theta}) \!=\! -\!\sum_{i q} a_i^q \delta_{v_i, q}\! -\! \sum_\mu b_\mu h_{\mu} \!-\! \sum_{i q \mu}\delta_{v_i, q}w_{i \mu }^q h_\mu,
\end{equation}
where $\delta_{v_i, q}$ is the Kronecker symbol that returns 1 if the site $i$ of the visible layer has the value $q$ and 0 otherwise, $\pmb{a} = \{a_i^q\}$ and $\pmb{b} = \{b_\mu\}$ are two sets of local fields acting respectively on the visible and on the hidden layers, and $\pmb{w} = \{w_{i \mu}^q\}$ is { a 3-rank tensor, which we simply refer to as} \textit{weight matrix}, that models the interactions between the two layers. We sketched the Potts-RBM structure in Fig. \ref{fig:Potts RBM}. { We would like to emphasize that the selection of Potts variables as visible variables for the RBM is a generalization that allows the study of sequence data within the same conceptual framework that applies to binary data. Nevertheless, binary Ising variables are sufficient for most clustering applications where the dataset can be made of binary states.}

The model defined by the Hamiltonian \eqref{eq:hamiltonian} is not uniquely defined, and common practice in Boltzmann learning is to fix the gauge in order to remove the flat directions caused by the over-parametrization~\cite{doi:10.1073/pnas.1111471108}. Recent works~\cite{tubiana:tel-02183417} suggested using the following Gauge fixing prescription
\begin{align}
    \sum_q a_i^q = 0, \qquad \sum_q w_{i \mu}^q = 0, \nonumber\\ \forall i = 1, \dots, N_v, ~\forall \mu = 1, \dots, N_h,
\end{align}
which is called \textit{zero-sum gauge}. {This choice preserves symmetry between the different categories and has the advantage of being computationally easier to implement. However, the choice of the actual gauge should not affect our clustering procedure.}

The RBM is trained by maximizing the Log-Likelihood (LL) function of the model computed on the dataset $\mathcal{D} = \{\pmb{v}^{(1)}, \dots, \pmb{v}^{(M)}\}$, which is defined as
\begin{align}\label{eq:LL}
    \mathcal{L}^{\mathrm{uw}}(\pmb{\theta} | \mathcal{D}) &=  \frac{1}{M} \sum_{m=1}^{M} \log p_{\pmb{\theta}}\left(\pmb{v} = \pmb{v}^{(m)}\right) = \nonumber\\ &= \frac{1}{M} \sum_{m=1}^{M} \log \sum_{\pmb{h}} e^{- E(\pmb{v}^{(m)}, \pmb{h}; \pmb{\theta})} - \log Z_{\pmb{\theta}},
\end{align}
via (stochastic) gradient ascent. For some datasets, however, it may be more convenient to look at a weighted version of the LL instead,
\begin{align}\label{eq:LL weighted}
    \mathcal{L}(\pmb{\theta} | \mathcal{D})\! &= \!\pa{\log p_{\pmb{\theta}}(\pmb{v})}_{\mathcal{D}} \!=\!  \frac{1}{M_{\mathrm{eff}}}\! \sum_{m=1}^{M} x_m \log p_{\pmb{\theta}}\left(\pmb{v}\! =\! \pmb{v}^{(m)}\right)\! \nonumber\\&=\! \frac{1}{M_{\mathrm{eff}}}\! \sum_{m=1}^{M} x_m\log \sum_{\pmb{h}} e^{- E(\pmb{v}^{(m)}, \pmb{h}; \pmb{\theta})} - \log Z_{\pmb{\theta}},
\end{align}
where $M_{\mathrm{eff}} = \sum_m x_m$ and $\{x_m\}_{m=1,\dots, M}$ is a set of weights associated with the data samples that have to be specified for each dataset. The weighted average likelihood is useful to compensate for the fact that certain categories may be overrepresented in the dataset compared to others. In particular, reweighting the dataset can affect the process of splitting the data into categories (especially since larger groups are split first in the tree), but has the disadvantage of requiring to know the groups in the first place. For this reason, we will generally not use this weighted version but for the protein sequence dataset to partially mitigate the fact that the data entries may not be independent of each other. This is typically the case for the available protein sequence data because the genomes of some species are sequenced more frequently than others~\cite{cocco2018inverse}. In this case, reweighting will not occur between functional groups but between sequences with similar sequence identity.

The gradient of the LL with respect to the model's parameters is given by
\begin{align}
    \frac{\partial \mathcal{L}}{\partial a_i^q} &= \pa{\delta_{v_i, q}}_{\mathcal{D}} - \pa{\delta_{v_i, q}}_{E}, \quad \frac{\partial \mathcal{L}}{\partial b_\mu} = \pa{h_\mu}_{\mathcal{D}} - \pa{h_\mu}_{E}, \nonumber\\ &\quad \text{and} \quad  \frac{\partial \mathcal{L}}{\partial w_{i \mu}^q} = \pa{\delta_{v_i, q} h_\mu}_{\mathcal{D}} - \pa{\delta_{v_i, q} h_\mu}_{E},
\end{align}
where we denoted as $\pa{\cdot}_{\mathcal{D}}$ the weighted average over the dataset and as $\pa{\cdot}_{E}$ the average over the model's Boltzmann measure \eqref{eq:Boltzmann distribution}. Since computing the partition function is generally intractable, this second average is typically approximated by the Block-Gibbs Sampling Markov Chain Monte Carlo (MCMC) method, which exploits the bipartite structure of the graph to iteratively sample the visible variables and the hidden variables conditioned on the other layer at once, hence allowing for a high degree of parallelization.
In particular, in this work we have always used the persistent contrastive divergence (PCD-$k$) method~\cite{tieleman2008training}: the initial states of the parallel Markov chains used to compute the gradient are taken from the final states obtained in the previous parameter's update. In this scheme, the number of steps performed in each update is fixed at $k$. It has been numerically investigated that this recipe can yield quasi-equilibrium models~\cite{decelle2021equilibrium}.
 We also systematically checked that the learned models always operated in the so-called equilibrium regime~\cite{agoritsas2023explaining,decelle2021equilibrium}, i.e., no memory of the training scheme was observed. In this paper, we will often refer to the number of epochs, $N_{\mathrm{epochs}}$, performed during the learning as the \textit{age} of the RBM, which can also be expressed in terms of the number of the gradient updates as $N_{\mathrm{grad}} = N_{\mathrm{epochs}} \cdot M/ N_{\mathrm{mb}}$, where $M$ is the number of training samples and $N_{\mathrm{mb}}$ is the minibatch size.

 \section{Using the RBM's learning history to classify the data}
 
\begin{figure*}[ht!]
    \centering
    \includegraphics[width=\textwidth]{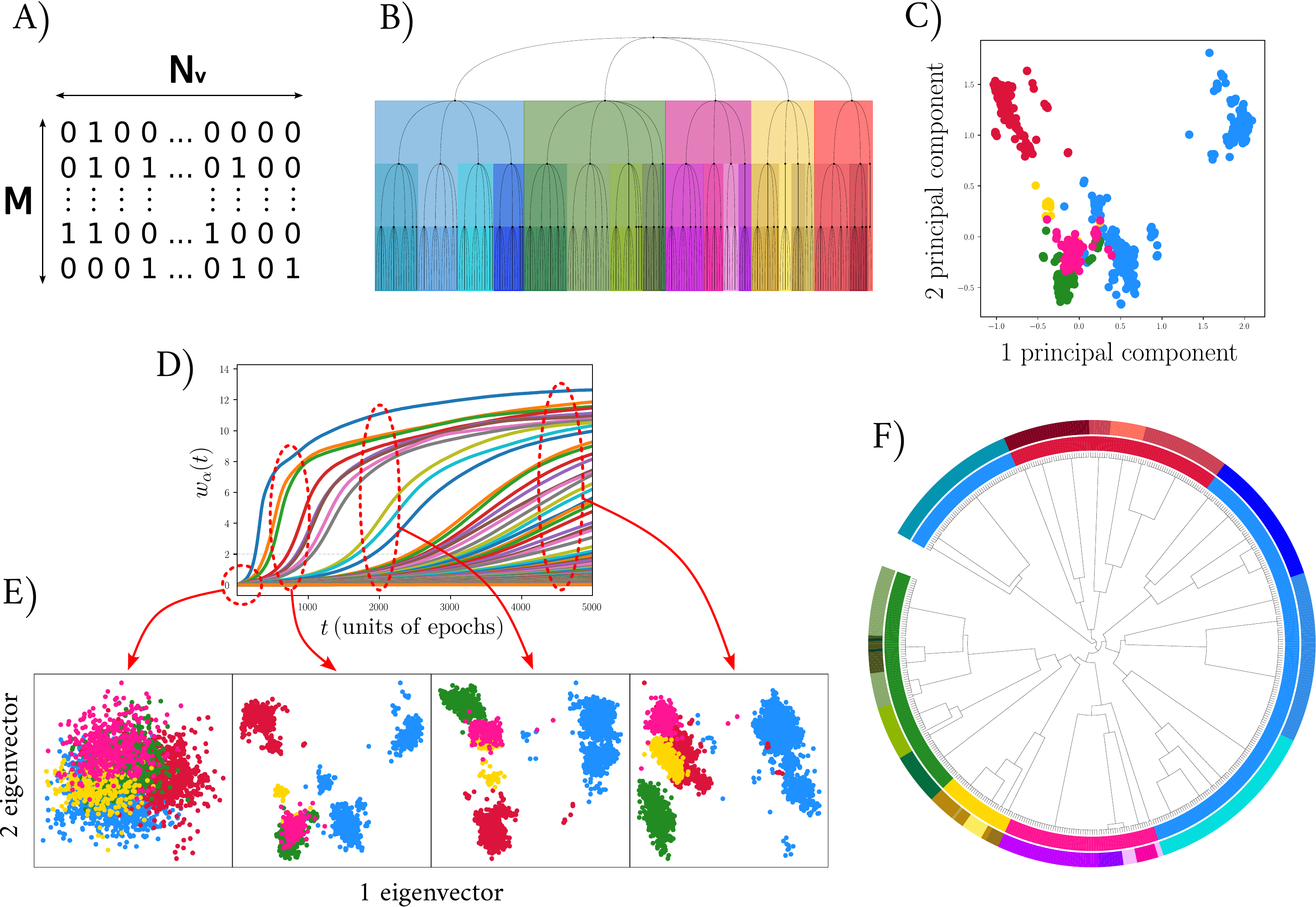}
    \caption{Different representations of the SED dataset and evolution of the features learned by the model as a function of the training time. A) SED dataset. Each row is a data vector made of $N_v$ binary values. B) First 4 levels of the real tree obtained from the SED dataset. Each node of the tree corresponds to a data sample. C) First 2 principal components of the SED dataset's PCA. D) Eigenvalues of the weight matrix, $\{w_{\alpha}\}$, as a function of the training time  in terms of epochs. E) Projections of the SED dataset along the first 2 principal directions of the weight matrix at different epochs. F) Tree of the SED dataset reconstructed with the proposed method. The coloured tags in the external rings represent the true labels (most internal ring) and sub-labels (outer ring) of the data points.}
    \label{fig:collage}
\end{figure*}

In this section, we discuss how the training trajectory of the RBM can provide useful information to reveal the hidden structure of a dataset. As a proof of concept, we artificially generated a dataset that mimics a completely random evolutionary process, which we call SED (Synthetic Evolutionary Data). We started with a ``mother sequence'' of zeros of length $N_v=805$ and generated a descendant set of 5 sequences by flipping each site from 0 to 1 with probability $p=1/N_v$. We then assigned a label to each of the child sequences and generated several more generations with the same criterion, each sequence having a randomly chosen number of heirs between 1 and 5. All descendants of any of the 5 ancestral sequences carry the same label and are referred to as a {\em family}. We can even go further by assigning a secondary label to each of the sub-trees originating from the grandchildren of the ``mother sequence''. Panels A and B of Fig.~\ref{fig:collage} show respectively a sketch of the structure of the binary dataset, and the genealogical tree corresponding to this evolutionary process. 

The most naive analysis we can do to infer families from the SED dataset is the PCA. We show the projections of the data onto the first 2 principal directions in Fig.~\ref{fig:collage}--C, and colour each data point according to its family colour. We can see that this analysis already does a relatively good job of separating sequences by family. For example, the red family (and perhaps the blue) can be directly isolated at this level of analysis, but the remaining points of the families remain mixed. Our goal now is to use the RBM to make a finer delineation and obtain some information about the internal structure of the families, e.g., to isolate the subfamilies of the grandchildren in Fig.~\ref{fig:collage}--B. { For the sake of visualization, the eigenvalues and eigenvectors of $\bm w$ are obtained after reshaping the tensor into the form of a $(N_v \cdot N_q) \times N_h$ matrix and taking its SVD (see Appendix~\ref{ap:proj} for details).} From previous works~\cite{decelle2017spectral,decelle2018thermodynamics,decelle2021restricted} we know that, during the initial steps of learning, the RBM gradually encodes the principal directions of the dataset in the SVD of the $\bm w$-matrix. In Fig.~\ref{fig:collage}--D we show the evolution of the eigenvalues of the weight matrix as a function of the training epochs, while in Fig.~\ref{fig:collage}--E we show the dataset projected along the first two eigenvectors of $\bm w$ at different stages of learning (different epochs). At the beginning of the learning process, the SVD eigenvectors do not contain useful information and cannot separate the families (all coloured points are mixed). In a second phase, when only the first 3 eigenvalues have grown, the projection closely resembles that of the PCA in Fig.~\ref{fig:collage}--C, thus agreeing with the previous statement that the SVD eigenvectors align with the direction of the principal directions~\cite{decelle2017spectral}. After that, the RBM starts learning nonlinear transformations of the data. In the third frame, corresponding to the point where the RBM has about 10 expressed eigenvalues, the projections partition the different families much better even at the level of a 2-dimensional projection. At much longer training times, we typically see that the clusters begin to overlap and direct clustering of such projections using a few dimensions becomes difficult, even when the machine has learned much better representations of the data.

With this discussion, we wanted to argue that the evolution of the free energy of the model during training can be used to explore different levels of clustering and in this way extract hierarchical information from the data. Such analysis can also give us qualitative insight into the feature extraction process in RBMs. Then, the underlying assumption is that the features learned by the RBM during the learning should have a hierarchical structure, that is, the ``most relevant'' properties of the data are learnt quite early during the training, while the finer substructure is detected only at later ages of the machine. The application of this method to the SED dataset yields the tree reported in Fig.~\ref{fig:collage}-F; we are going to discuss it in the next sections. We see that data are correctly separated into distinct families and sub-families and a rich structure of the dataset has been detected, although the blueish data get split into two distinct branches right from the beginning of the tree. The criterium we used for categorizing the data at different ages of the learning is based on the mean-field theory applied to the RBM, and it is discussed in the next section.

\section{Probing the RBM landscape: a mean-field analysis}

The method that we propose is based on the mean-field description of the RBM.
In this context, we can describe the system through an approximated thermodynamic potential which depends on the system's magnetizations. This extended free energy displays a certain number of metastable minima in the magnetization space~\cite{zamponi2010mean}, each of which corresponds to a mode (i.e., a local maximum) of the Boltzmann probability distribution. 
We can interpret these metastable states as \textit{features} of the data that the model was able to capture at a certain stage of the learning.
Using the mean-field theory, we can derive a set of self-consistent equations that, iterated until convergence, allow us to associate each data point in the dataset to the ``closest'' metastable state, thereby categorizing/clustering the data into families with a similar structure to the RBM eyes.
Thus, by following this assignment procedure during training for different free energy landscapes of progressively trained machines, we have a method for hierarchically classifying the data.

\subsection{Derivation of the self-consistent equations for the Potts RBM}
In what follows, we generalize the mean-field description of Ref.~\cite{Plefka_1982,georges1991expand,tanaka2000information}, also known as Plefka expansion, to our target model defined by the Hamiltonian in equation~\eqref{eq:hamiltonian}. These high-temperature expansions have been already applied to the Boltzmann Machine~\cite{tanaka1998mean} and to RBMs { with binary visible variables}~\cite{gabrie2015training,tramel2018deterministic,maillard2019high} and to the infinite-range Potts-glass model~\cite{PhysRevB.51.962}, but we have yet to generalize these results to the RBM case with Potts variables in the visible layer. In this section, we outline only the main results that will be used later to build the trees, while further details on the derivation can be found in the supplementary text~\ref{app:TAP equations for the Potts RBM}.

Let us consider an energy function $E_\beta(\pmb{v}, \pmb{h}; \pmb{\theta})$ which depends on the inverse-temperature $\beta$. By introducing two sets of temperature-dependent external fields $\pmb{B} \!=\! \{B_i\}\!=\! (\{\phi_j^p\},\{\psi_\mu\})$ with $i \in \{1, \dots, N_v \cdot N_q + N_h\}$, interacting respectively with the visible and the hidden variables, we consider the free energy of this extended system:
\begin{align}
    - \beta F_{\beta}(\pmb{B}) =& \log \sum_{\pmb{v}, \pmb{h}} \exp\biggl[-\beta E_\beta(\pmb{v}, \pmb{h}; \pmb{\theta}) + \nonumber\\ &+ \sum_{i q} \phi_i^q(\beta) \delta_{v_i, q} + \sum_\mu \psi_\mu(\beta) h_\mu\biggr],
    \label{eq:free energy extended system MT}
\end{align}
where we omit the dependence of the free energy on the systems' parameters $\pmb{\theta}$. By taking the Legendre transform of \eqref{eq:free energy extended system MT}, we obtain a description of the system in terms of the magnetizations $\pmb{M} = \{M_i\} = (\{f_j^p\},\{m_\mu\})$ through the Gibbs free energy
\begin{widetext}
\begin{align}
    -\beta G_\beta(\pmb{M}) &= -\beta \max_{\pmb{B}} \biggl[F_\beta(\pmb{B}) + \frac{1}{\beta} \sum_{i q} \phi_i^q(\beta) f_i^q +\frac{1}{\beta} \sum_\mu \psi_\mu(\beta) m_\mu\biggr] = \nonumber \\
    &= \min_{\pmb{B}} \biggl[\log \sum_{\pmb{v}, \pmb{h}} \exp \biggl(-\beta E_\beta(\pmb{v}, \pmb{h}; \pmb{\theta}) + \sum_{i q} \phi_i^q(\beta)(\delta_{v_i, q} - f_i^q) + \sum_\mu \psi_\mu(\beta)(h_\mu - m_\mu)\biggr)\biggr],
    \label{eq:Gibbs free energy MT}
\end{align}
\end{widetext}
where the maximum condition is taken because $F_\beta(\pmb{B})$ is a concave function of the fields. Consequently, this new thermodynamic potential is a convex function of the magnetizations.
For the properties of the Legendre transform, fields and magnetizations are related through
\begin{align}
    M_i(\pmb{B}) = -\beta \frac{\partial}{\partial B_i} F_\beta(\pmb{B}) \quad \text{and} \quad B_i(\pmb{M}) = \beta \frac{\partial}{\partial M_i} G_\beta(\pmb{M}).
\end{align}
The extended free energy \eqref{eq:free energy extended system MT} can be then recovered as the inverse Legendre transform of the Gibbs free energy \eqref{eq:Gibbs free energy MT}
\begin{align}
    -\beta F_\beta(\pmb{B}) =& -\beta \min_{\pmb{M}} \biggl[G_\beta(\pmb{M}) - \frac{1}{\beta} \sum_{i q} \phi_i^q(\beta) f_i^q + \nonumber\\ &- \frac{1}{\beta} \sum_\mu \psi_\mu(\beta) m_\mu\biggr].
    \label{eq:inverse Legendre transform MT}
\end{align}
From equation \eqref{eq:inverse Legendre transform MT}, we see that the free energy of the original system -- corresponding to the free energy \eqref{eq:free energy extended system MT} when the external fields are turned off -- can be obtained as the minimum of the Gibbs free energy
\begin{align}
    A_\beta &= -\frac{1}{\beta} \log \sum_{\pmb{v}, \pmb{h}} e^{-\beta E_\beta(\pmb{v}, \pmb{h}, \pmb{\theta})} =\nonumber\\ &= F_\beta(\pmb{B} = \pmb{0}) = \min_{\pmb{M}} \pq{G_\beta(\pmb{M})}.
    \label{eq:free energy stationary conditions MT}
\end{align}
In this way, we derived a method to compute the free energy of the system by operating in the space of magnetizations. However, this is true only in principle, because the Gibbs free energy is as difficult to compute as the true free energy $A_\beta$. Besides, our goal of detecting the metastable states of the system cannot be accomplished by working with the Gibbs free energy, because it displays just one global minimum. An estimation of the Gibbs free energy and the breaking of the minima degeneracy can be achieved by considering a high-temperature regime in which we can approximate the thermodynamic potential with a Taylor expansion around $\beta = 0$ truncated at a certain order~\cite{zamponi2010mean}. In practice, the high-temperature limit corresponds to a regime in which interactions among the variables are negligible compared to thermal fluctuations. This is the typical situation that we encounter at the beginning of the learning, when the weight matrix stores only minimal information on how to couple the layers of the RBM to reproduce the dataset's statistics. To formalize this, we can consider a Hamiltonian of the kind
\begin{equation}
    -\beta E_\beta(\pmb{v}, \pmb{h}; \pmb{\theta}) = \sum_{i q} a_i^q \delta_{v_i, q} + \sum_\mu b_\mu h_{\mu} +\beta \sum_{i q \mu}\delta_{v_i, q}w_{i \mu }^q h_\mu,
\end{equation}
where the inverse temperature $\beta$ is just a dummy parameter that serves for the Taylor expansion and that we will eventually set to 1. Indeed, by re-defining $\Tilde{\pmb{w}} = \beta \pmb{w}$ we see that the limit $\beta \rightarrow 0$ is equivalent to the limit $\Tilde{w}_\alpha \rightarrow 0$, being $\Tilde{w}_\alpha$ the eigenvalues of the weight matrix. That is indeed the case when the RBM has not yet learned the modes of the dataset~\cite{decelle2017spectral}.

By defining $\Gamma_\beta(\pmb{M}) = -\beta G_\beta(\pmb{M})$ for simplicity, we obtain the following second-order expansion of the Gibbs free energy:
\begin{widetext}
\begin{alignat}{3}
    \Gamma_\beta^{(2)}(\pmb{M}) &= &&\Gamma_0(\pmb{M}) + \beta \evalat{\frac{\partial \Gamma_\beta(\pmb{M})}{\partial \beta }}{\beta=0} + \frac{\beta^2}{2} \evalat{\frac{\partial^2 \Gamma_\beta(\pmb{M})}{\partial \beta^2}}{\beta=0} =\nonumber \\
    &= && \sum_{i q} f_i^q a_i^q + \sum_\mu m_\mu b_\mu - \sum_{i q} f_i^q \log f_i^q - \sum_\mu \pq{m_\mu \log m_\mu + (1 - m_\mu) \log (1-m_\mu)} + \nonumber \\
    & &&+ \beta \sum_{i q \mu} f_i^q w_{i \mu}^q m_\mu + \frac{\beta^2}{2} \sum_\mu (m_\mu - m_\mu^2) \pq{\sum_{i q} (w_{i \mu}^q)^2  f_i^q - \sum_{i} \pt{\sum_q w_{i \mu}^q f_i^q}^2}.
    \label{eq:TAP Gibbs free energy MT}
\end{alignat}
\end{widetext}
The stationary conditions \eqref{eq:free energy stationary conditions MT} yield a pair of self-consistent equations which are known as the Thouless-Anderson-Palmer (TAP) equations~\cite{thouless1977solution} and are the equivalent of the ones already obtained for RBM with different input types in Refs.~\cite{gabrie2015training,tramel2018deterministic}
\begin{widetext}
\begin{align}
\begin{split}
    m_\mu &= \mathrm{sigmoid}\pg{b_\mu + \sum_{i q} f_i^q w_{i \mu}^q - \pt{m_\mu - \frac{1}{2}} \pq{\sum_{i q} (w_{i \mu}^q)^2  f_i^q - \sum_{i} \pt{\sum_q w_{i \mu}^q f_i^q}^2}},\\
    f_i^q &= \mathrm{softmax}\pg{a_i^q + \sum_\mu m_\mu w_{i \mu}^q + \frac{1}{2}\sum_\mu (m_\mu - m_\mu^2) \pq{(w_{i \mu}^q)^2  - 2 w_{i \mu}^q \sum_{p}f_i^p w_{i \mu}^p}}.
    \label{eq:TAP equations}
\end{split}
\end{align}
\end{widetext}

{
The approximated free energy \eqref{eq:TAP Gibbs free energy MT} has many local minima, unlike the full Gibbs free energy of Eq.~\eqref{eq:Gibbs free energy MT}, which is convex. Although this fact could be considered a pure artefact of high-temperature expansion, in practice these various metastable states can be associated with relative maxima in the probability distribution~\cite{zamponi2010mean}. This fact can be explicitly verified by direct visual inspection of a dimension reduction of the generated data and the fixed points, at least at early steps in the learning history when the generated data are highly clustered. Moreover, one can also verify that this multiplication of local minima does not vanish when we include higher orders of $\beta$ in the expansion. In particular, we have systematically compared the fixed points obtained with the third order of the Plefka expansion~\cite{gabrie2015training} with those obtained from the TAP equations in the binary RBM case for some specific data sets, and we have not found any appreciable difference that would justify the implementation of higher orders than the second in the free-energy expansion.} 

{ We want also to stress that there are other methods that allow approximating the free energy of the model through high-temperature expansions to higher orders than the second, such as Belief Propagation (BP) \cite{huang2021statistical}. However, these methods suffer from the same main limitation of the Plefka expansion, for instance, the BP algorithm assumes the weights of the network to be uncorrelated from each other, additionally approximating the factor graph of the model as a tree. We know that this assumption is only true at the early stages of the learning trajectory when the weights are small~\cite{decelle2017spectral,decelle2021restricted}.}

{ One could in principle use these TAP fixed points as a way to speed up the training procedure, as suggested in previous works~\cite{gabrie2015training,tramel2018deterministic}. While this procedure is very effective for training models with good generative performance, there is no guarantee that the Boltzmann distribution of these models is an accurate representation of the dataset~\cite{agoritsas2023explaining}. In fact, for this work, we investigated this training procedure and obtained very poor results when we tried to use the RBMs trained with the TAP equations for clustering objectives. We will discuss these results later.
}

\subsection{Clustering data using the TAP fixed points}
\begin{figure*}[ht!]
    \centering
    \includegraphics[width=\textwidth]{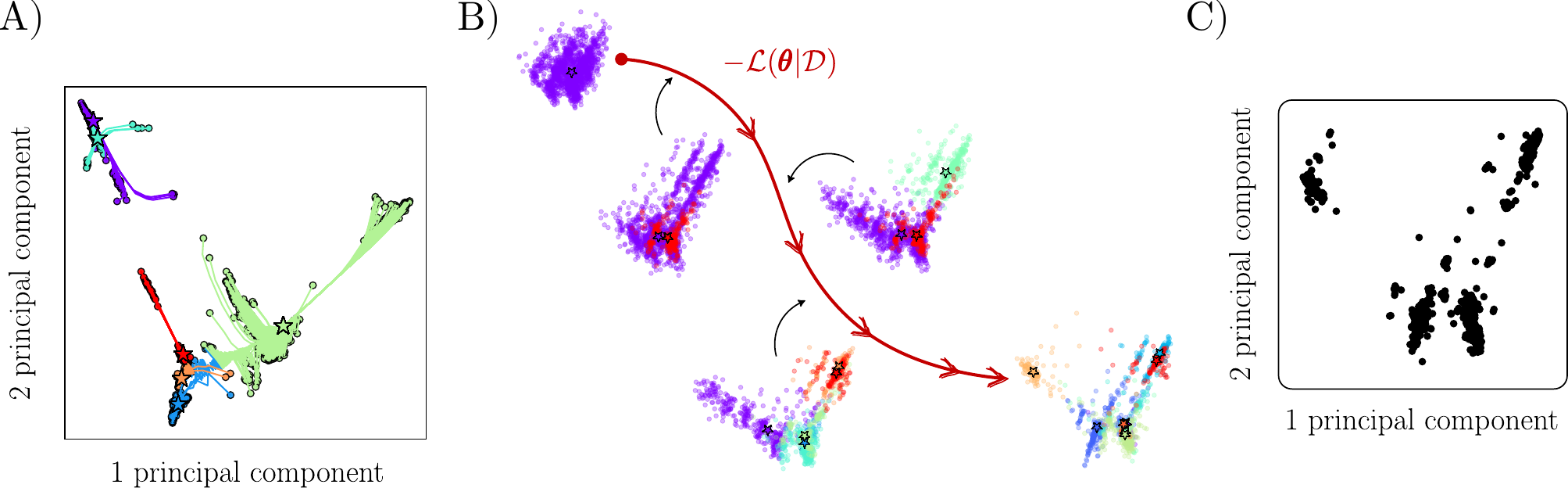}
    \caption{A) Representation of the TAP equations' dynamics projected along the two main components of the PCA. The coloured circles represent the initial conditions (magnetizations) of the TAP equations associated with the data points of the SED dataset. The coloured lines describe the trajectories that the points follow when iterating the TAP equations, while the stars represent the final fixed points. The colours of circles and lines match that of the corresponding fixed points. B) Pictorial representation of the (minus) LL's minimization trajectory during the learning on the SED dataset. The scatter plots represent the data generated by the model at different ages projected along the first two components of the training dataset's PCA. The coloured stars correspond to the fixed points of the TAP equations initialized with the data, and the data points' colour matches the colour of the corresponding fixed point. Notice how, as the training advances, the generated dataset becomes more and more structured, and the number of TAP fixed points increases accordingly. C) $\sim 500$ samples from the SED training dataset projected on the first two principal components of the PCA. Notice how the picture resembles the projection of the last generated dataset on panel B.}
    \label{fig:LL trajectory}
\end{figure*}

{
Given the model's parameters at a certain age, we can use the dataset as initial conditions by mapping each data point $\pmb v^{(n)}$ to a pair of magnetization sets $(\{f_i^{q,(n)}(t=0)\}, \{m_\mu^{(n)}(t=0)\})$ using
\begin{align}
    m_\mu^{(n)}(t=0) &= \mathrm{sigmoid}\pt{b_\mu + \sum_{iq}\delta_{v_i^{(n)}, q} w_{i \mu}^q}, \nonumber \\
    f_i^{q,(n)}(t=0) &= \mathrm{softmax}\pt{a_i^q + \sum_{\mu} m_\mu^{(n)}(t=0) w_{i \mu}^q}
\end{align}
and then iterate equations \eqref{eq:TAP equations} until convergence.} In this way, the data belonging to the same basin of attraction of the free energy will collapse into the same point of the magnetization space, corresponding to one of the possibly many local minima, as shown in Fig. \ref{fig:LL trajectory}-A. We can then use these multiple basins of attraction to cluster similar data together. We just need a practical way to automatically collect all the data in the same basin (or metastable state).

To this end, we can use the DBSCAN algorithm~\cite{10.5555/3001460.3001507} to categorize the data by the fixed point at which it ends after propagating the TAP equations. Now, if we do this at different stages of learning, such a method allows us to partition the data set according to the features that the model has learned at that point (age) of training. This method is quite robust in choosing the \textit{eps} parameter of DBSCAN, because the TAP dynamics makes the points collapse into very small regions of magnetization space that can be easily delineated by the algorithm without ambiguity. However, a relatively large value of \textit{eps}=1 was chosen because we sometimes want to aggregate fixed points that almost overlap but are still distinct, preferring larger categories than having extremely similar fixed points. In Fig. \ref{fig:LL trajectory}-B, we sketch how the number of fixed points and the features recognized by the model evolve during the learning. Data points that converge to the same fixed point (represented by a star) are coloured accordingly. This construction allows us to visualize how the free energy landscape changes during the training, displaying a growing number of local minima. We note that this procedure is very similar to the annealing performed on the Gaussian mixture model~\cite{PhysRevLett.65.945,PhysRevE.103.012105}.

We remark that the fixed points of the TAP equations are not guaranteed to be local minima, but they might be saddle points or local maxima as well. 
In principle, this could be verified by computing the determinant Hessian matrix of the TAP free energy \eqref{eq:TAP Gibbs free energy MT} at the fixed points. Yet, in practice, this check reveals quite problematic since one should compute a $(N_v \cdot N_q + N_h) \times (N_v \cdot N_q + N_h)$ dimension matrix for each one of the fixed points that are identified. Instead, we qualitatively verified that the fixed points in the early phase of learning always correspond to the local minima of free energy by visually verifying that they fall in the middle of the denser regions, i.e., identifying the modes of the probability distribution. However, in the oldest machines, we can not guarantee that some of the fixed points may be spurious solutions. To minimize this problem, we will always take a ``backward'' approach to the tree construction, as will be explained in the next section. 

\begin{figure*}[t!]
    \centering
    \includegraphics[width=0.7\textwidth]{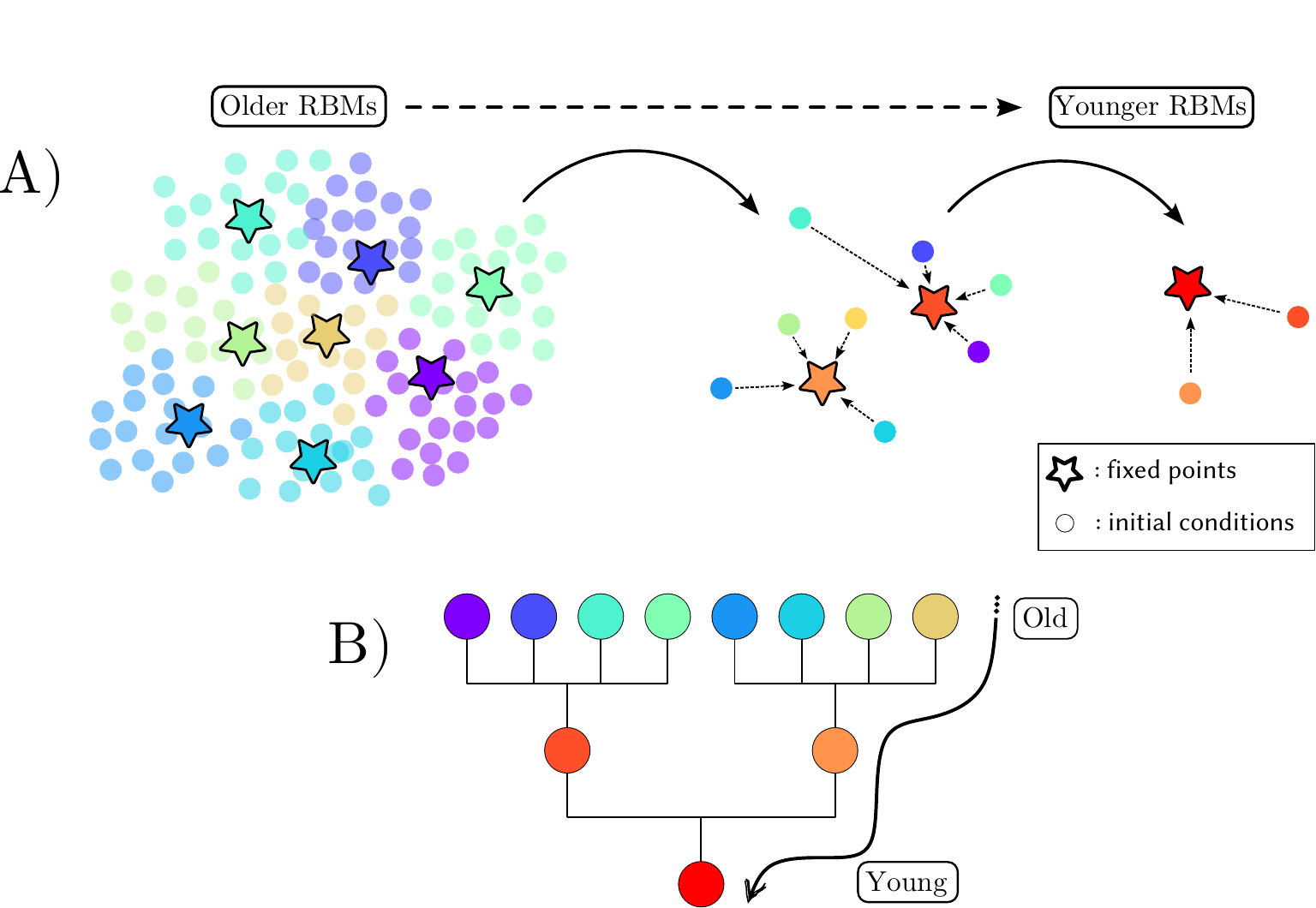}
    \caption{A) Sketch of the process of merging the fixed points used to build the relational tree. We start with an old RBM and let the TAP dynamics collapse the data into the fixed points of the self-consistent equations \eqref{eq:TAP equations}. Then we repeat the procedure for younger and younger machines, each time propagating the TAP equations starting from the previously found fixed points, until we reach the point where all the data have collapsed into just one global minimum. B) By following the merging of the fixed points, we obtain the relational tree of the dataset.}
    \label{fig:tree construction}
\end{figure*}

\section{Generating relational trees}
In this section, we discuss how to use the aforementioned construction to produce a hierarchical description of the dataset.
The method we propose consists of considering multiple classifications performed using several models saved at different stages of the training (i.e. \textit{ages}). Suppose to have an ordered set of ages $\{t_{\mathrm{age}}^{k}\}_{k=1, \dots, N_{\mathrm{T}}}$ with $t_{\mathrm{age}}^{k+1} > t_{\mathrm{age}}^{k}$. We start by taking the oldest RBM at $t_{\mathrm{age}}^{N_{\mathrm{T}}}$ and we iterate the TAP equations until convergence using the dataset as initial conditions. This procedure yields a number of fixed points which is smaller than the number of data points since some data points will have collapsed into the same free energy minimum. Then, we pick the model at $t_{\mathrm{age}}^{N_{\mathrm{T}}-1}$ and we iterate again the TAP equations, this time using as initial conditions the fixed points found at the previous step. In fact, some minima that were distinct for the oldest RBM will be degenerate for the younger one, finding themselves in the same basin of attraction of a new (larger) minimum. We can repeat this procedure backwards in time for all the selected ages until the youngest RBM, for which all the data collapse into a unique global minimum. If we keep track of the fixed points' merging during the procedure, we can construct a tree starting from the leaves (i.e., splits made by old models) and proceeding down to the root. We depicted this procedure in Fig.~\ref{fig:tree construction}.
 This procedure also allows us to efficiently avoid the spurious fixed points (the saddle points or maxima of free energy): we follow the merging of fixed points obtained in old machines into the broader basins of attraction of the fixed points of the younger machines. This procedure ensures that the tree construction will rapidly rely only on the true minima of the free energy. Yet, one could also imagine constructing a tree based on the fixed points that arise when propagating the TAP equations from all points in the data set at each age of the machine. We have found empirically that trees constructed in this way are qualitatively and quantitatively worse than those presented here (see Fig.~\ref{fig:tree MNIST wrong way} of the supplementary information). We believe that a reason for this is precisely that not all fixed points are free-energy minima.
Another possible explanation for this is that, in this second case, classifications performed at different ages are independent of each other. In the method we described, instead, the classification at a given age is informed by those performed at later ages. In other words, information is propagated backwards in (training) time, from the more experienced models to the rawest ones.

\begin{figure*}[ht!]
    \centering
    \includegraphics[width=0.7\textwidth]{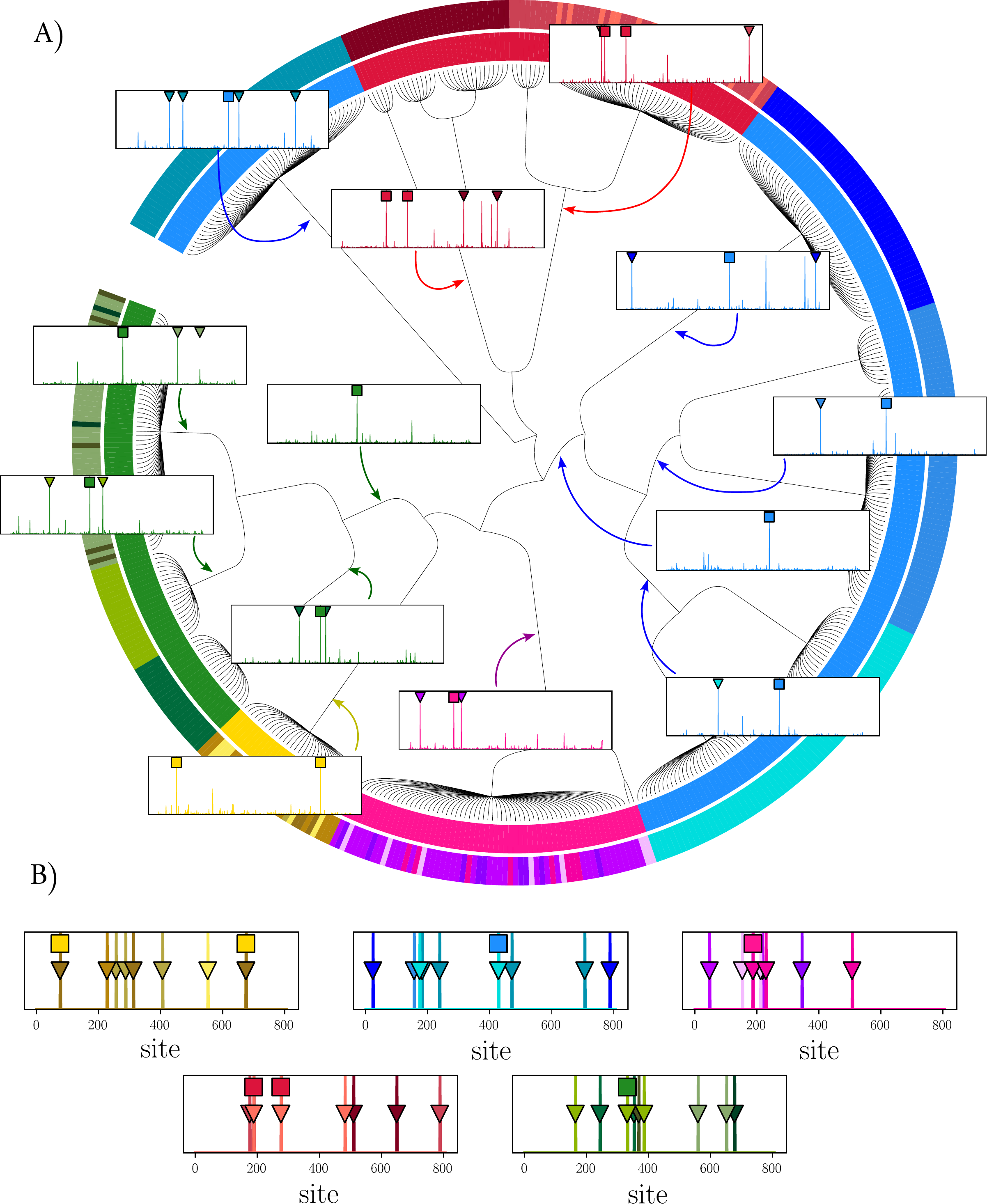}
    \caption{Representation of the features learned by the RBM during training on the SED dataset. A) The same tree as in Fig.~\ref{fig:collage}-F, truncated to 15 layers. The plots in the insets represent the comparison between the magnetization profiles (solid colored lines) associated with some internal nodes of the tree and the ``patterns" of the first and second generations of the genealogical tree in Fig.~\ref{fig:collage}-B. The x-axis of the boxes denotes the index site along the sequence, whereas the y-axis represents the value of the magnetization. The coloured squares indicate the locations where mutations occurred in the first generation of progeny (passing from the ancestral mother sequence with all 0s mother sequence, to a sequence of 0s plus some few 1s), and the triangles indicate the mutations in the second generations corresponding to this subset (secondary colouring labels). B) Representation of the location of all first and second-generation mutations. Each box represents one of the first 5 families. The squares mark the mutations in the first generation, while the triangles and the vertical lines highlight the mutation sites in the second generation (in both cases compared to the ancestral mother sequence). Each colour represents a different second-generation family (retaining the colour scheme of Fig.~\ref{fig:collage}--B).}
    \label{fig:SEDtreefeatures}
\end{figure*}

At this point, we have to state a criterion for selecting the ages of the RBM that we want to save during the training and that will participate in the tree construction. Ideally, one would wish to find a relation between the geometry of the free energy landscape (e.g. the number of fixed points that are present) and some observable that can be monitored during the training, such as the spectra of the weight matrix, to decide optimally and automatically when to save the state of the RBM. The first splits are easily traceable by monitoring the first eigenvalues of the weight matrix. Yet, when the RBM starts to have an increasing number of ``expressed'' modes, we can not anymore link one mode to a given split (see Fig.~\ref{fig:eig vs TAP} of the supplementary material).
For this reason, we simply saved a certain number of models (e.g. $\sim 500$) equally spaced along the training history. When constructing the tree, we added a new layer only when the number of fixed points found with the TAP equations decreased (recall that we move backwards in time when using the TAP equations) from the previously added layer.

Applying this procedure to the artificial dataset SED, we create the tree we showed as teasing in Fig.~\ref{fig:collage}--F. The leaves of the graph (real data points) were colored according to the labels of each entry, as shown in the genealogical tree in Fig.~\ref{fig:collage}--B. We consider two possible levels of relationships: Ancestral families are shown in 5 colours (blue, green, pink, yellow, and red) and next-generation entries are shown in different shades of the same color family. We see that with this tree we can perfectly delineate all 5 families based on the main branches of the constructed tree, and the same is true for the subfamilies initiated by the grandchildren (shown in the outer crowns of the diagram). Since the nodes of the tree correspond to points in the magnetization space, we can visualize the evolution of the learned features during learning, as shown in Fig.~\ref{fig:SEDtreefeatures}--A. Looking at the innermost nodes of the tree, we can see that the RBM first learns the characteristic patterns of the first generation of sequences by recognizing the mutation sites, and then gradually distinguishes the second-generation mutations, and so on. Interestingly, the magnetizations of the various fixed points tell us exactly where the mutations (which led to the creation of each family and subfamily) occurred during the evolutionary process (see Fig.~\ref{fig:SEDtreefeatures}--B).
Furthermore, from this diagram, it is easy to  understand why the main blue family is already divided into two main groups at the root of the tree. The magnetization profile of the blue family in the upper left part of Fig.~\ref{fig:SEDtreefeatures}, indeed, shows a burst of mutations that occurred very early in the evolutionary process and caused this subfamily to diverge from the rest of the sequences.

\section{Results}

So far, we have shown the performance of our method only on a synthetic dataset whose hierarchical relational structure was given by design. Now, we apply our method to 3 different real-world datasets with different categories of data points. First, we analyze a simple image dataset where all labels are known and where features can be visualized directly. In particular, we analyze a relational structure of the MNIST dataset~\cite{lecun1998gradient}, which contains images of handwritten digits whose label is the actual digit. Second, we analyze the Human Genome Dataset (HGD)\cite{10002015global}, which captures the genetic mutations of 5008 individuals from around the world. This dataset is even more interesting because the sequences contain two levels of annotation (continental origin and population group). Third, we conclude with the last and most challenging dataset: amino acid sequences of a homologous family of proteins (the CPF protein) with a large functional diversity. The biological function of these sequences is known for only a very small fraction, namely those that have been studied experimentally. The challenge, then, is to assign a probable biological function to each of the unclassified sequences based on limited experimental knowledge. For a detailed description of the datasets considered, see section~\ref{sec:datasets} in the Supplementary Material.

\begin{figure*}[ht!]
    \centering
    \includegraphics[width=\textwidth]{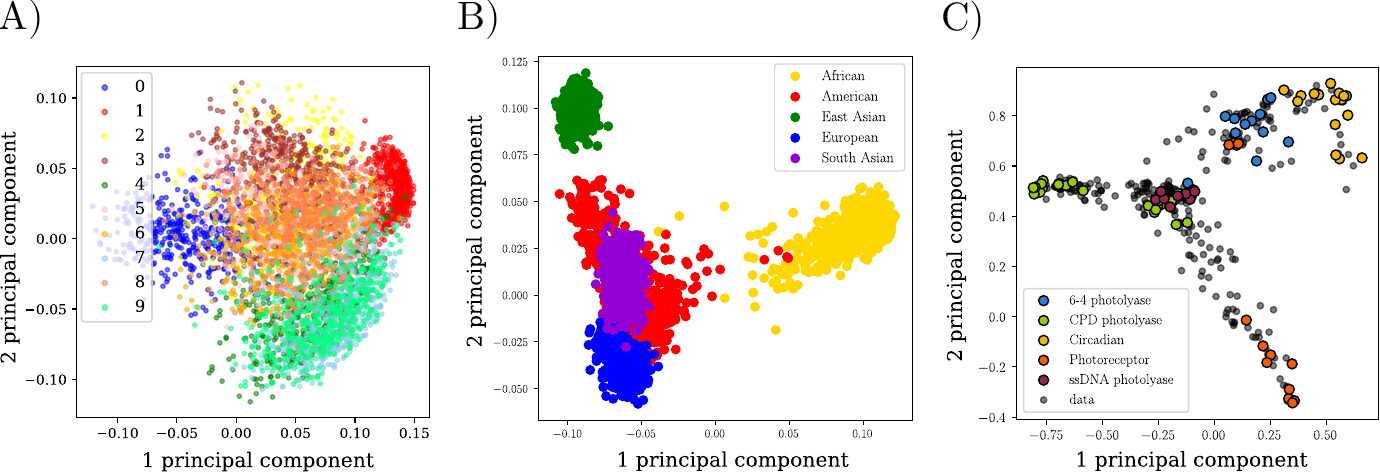}
    \caption{Projections of the datasets along the first two principal components of the PCA. The colours correspond to the known labels of the data points. A) MNIST dataset. B) Human Genome dataset. C) CPF protein. All the data are shown in black and the coloured dots correspond to the experimentally annotated sequences. }
    \label{fig:PCA all}
\end{figure*}

We show the projection of all these 3 datasets along the first two principal  components of the PCA in Fig.~\ref{fig:PCA all}. In particular, we point out the different degree of clusterization displayed in each of these 3 datasets: while the distribution of the MNIST data points is quite broad and connected, the HGD and the CPF protein are rather clusterized, which is known to sensibly damage the quality of the training process due to sampling convergence problems~\cite{10.21468/SciPostPhys.14.3.032}.

\subsection{MNIST dataset}

\begin{figure*}[ht!]
    \centering
    \includegraphics[width=0.8\textwidth]{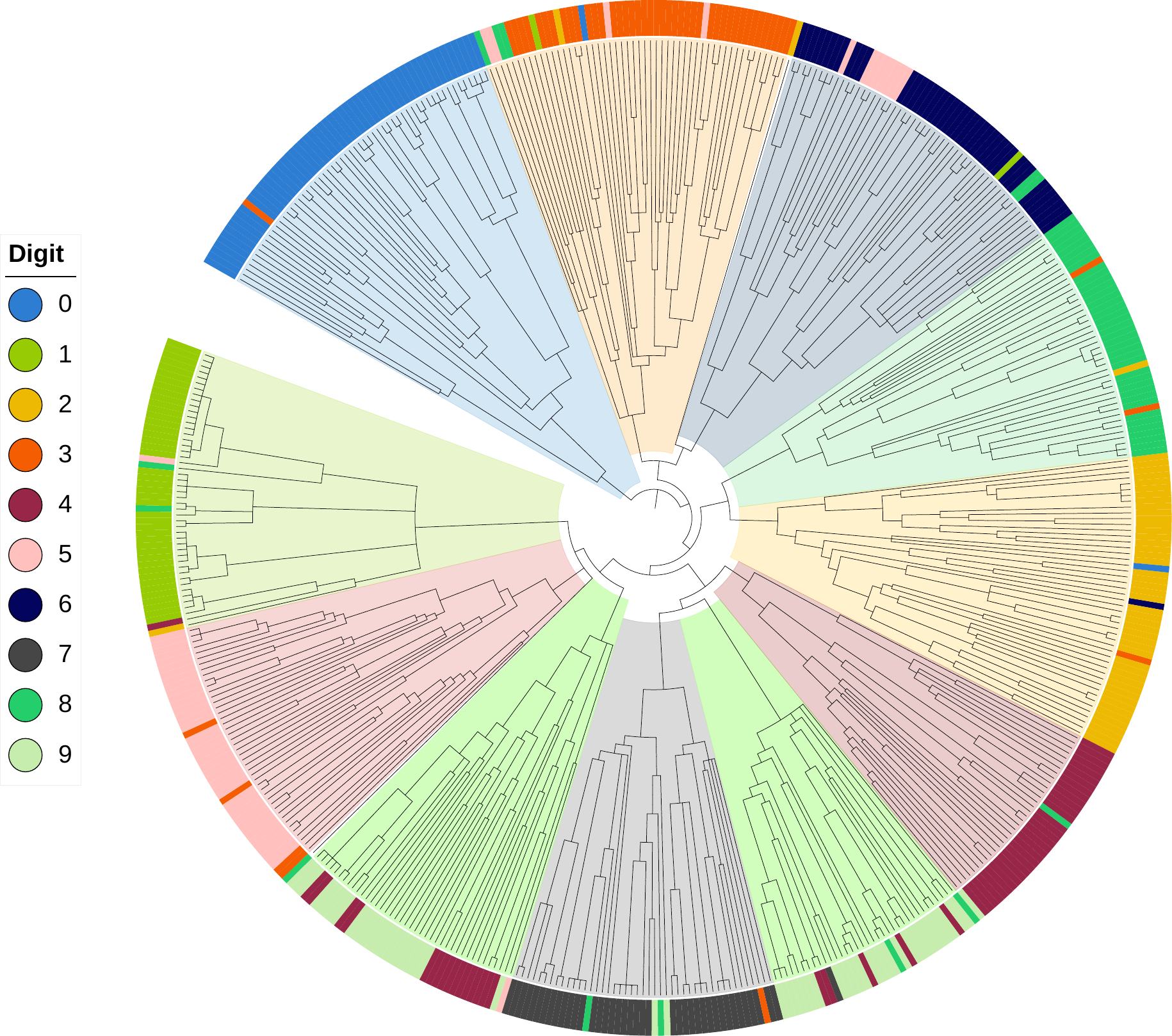}
    \caption{Complete tree generated on 500 samples from the MNIST dataset. The external coloured tags represent the true labels, while the coloured regions inside the tree have been manually assigned for the sake of visualization.}
    \label{fig:tree MNIST}
\end{figure*}

\begin{figure*}[ht!]
    \centering
    \includegraphics[width=0.8\textwidth]{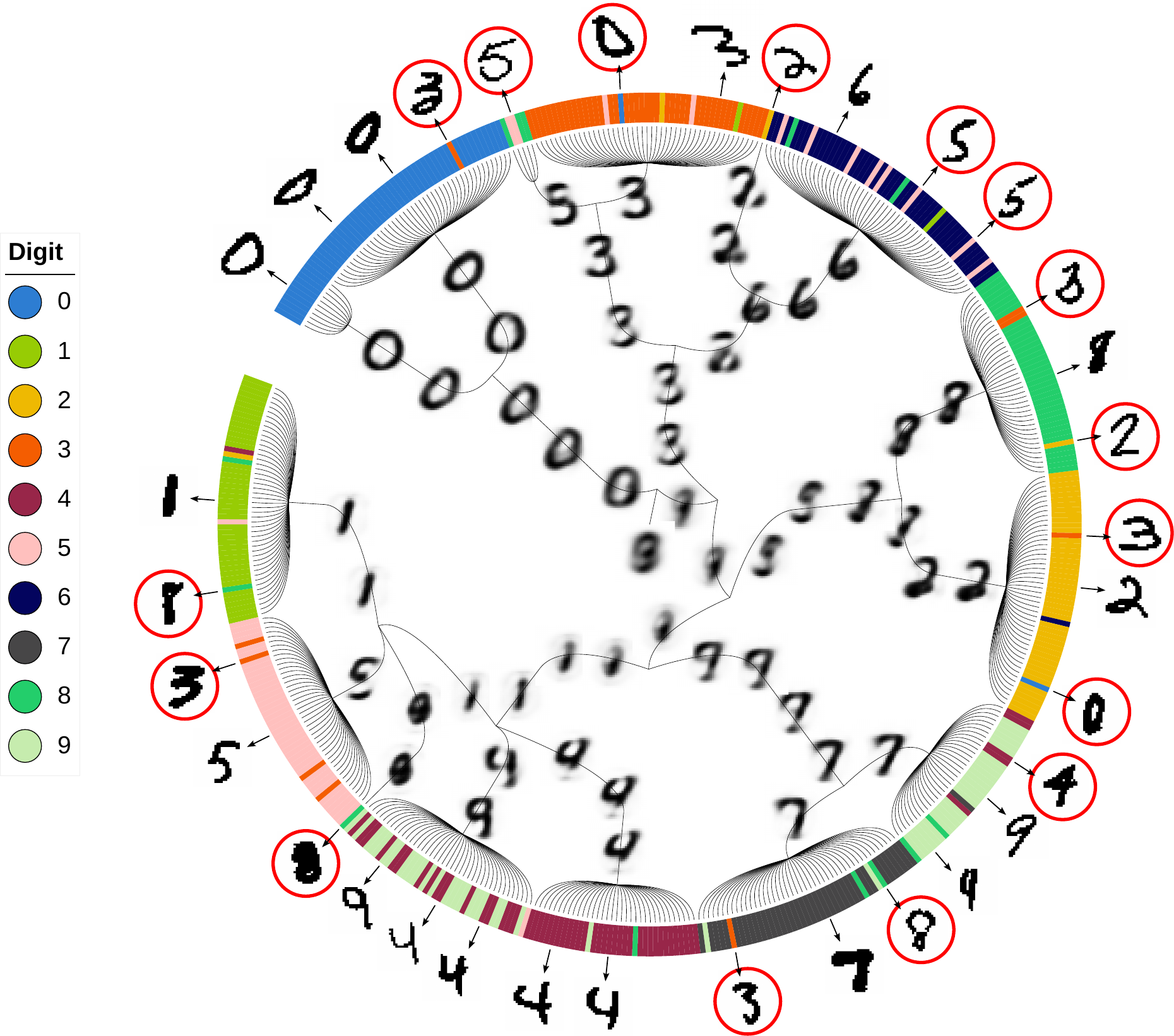}
    \caption{First 10 layers of the tree obtained on 500 samples from the MNIST dataset. The coloured tags in the external ring correspond to the true labels of the data. Inside the three, we superimposed the images of the RBM's fixed points corresponding to the nodes along the branches. The intensity of the pixels represents the probability that the machine assigns to that pixel to be turned on. We also report the images from the dataset corresponding to some of the leaves. In particular, we put a red circle around the digits that have been wrongly classified by the algorithm in order to highlight the configurations that fooled the machine.}
    \label{fig:tree images MNIST}
\end{figure*}

The MNIST dataset consists of binary images (pixels can be either 0 or 1) of handwritten digits. Each example is associated with a numeric label - from 0 to 9 - indicating the digit represented in the image. We show in Fig.~\ref{fig:tree MNIST} the relational tree obtained by our method for MNIST. As we can see, the RBM is mostly able to separate the different digits, although there are some cases where two categories are mixed together (e.g., the digit 4 with the 9). Dealing with pictures has the advantage that we can visualize the features learned by the RBM during learning and try to explain them qualitatively.
The first few layers of the tree obtained using a subset of the MNIST dataset are shown in Fig.~\ref{fig:tree images MNIST}, where we also highlighted the images of some data that were misclassified. Since the internal nodes of the tree correspond to actual points in the magnetization space, we can visualize the evolution of the features learned from the RBM as they unfold along the tree and see where and when each cluster detail appears.
For example, the first split separates digit 0 from the rest. As training time (or tree depth) increases, the 0 appears to remain completely isolated while the rest of the digits are gradually resolved as the RBM learns finer details. We can also observe the proximity of some digits; in addition to the 4 and the 9, we see that the 7 is also quite close to the 9. In principle, it would be possible to further analyze the obtained results and find interesting features that may have been unknown before the analysis.

\subsection{Human Genome dataset}

\begin{figure*}[ht]
    \centering
    \includegraphics[width=\textwidth]{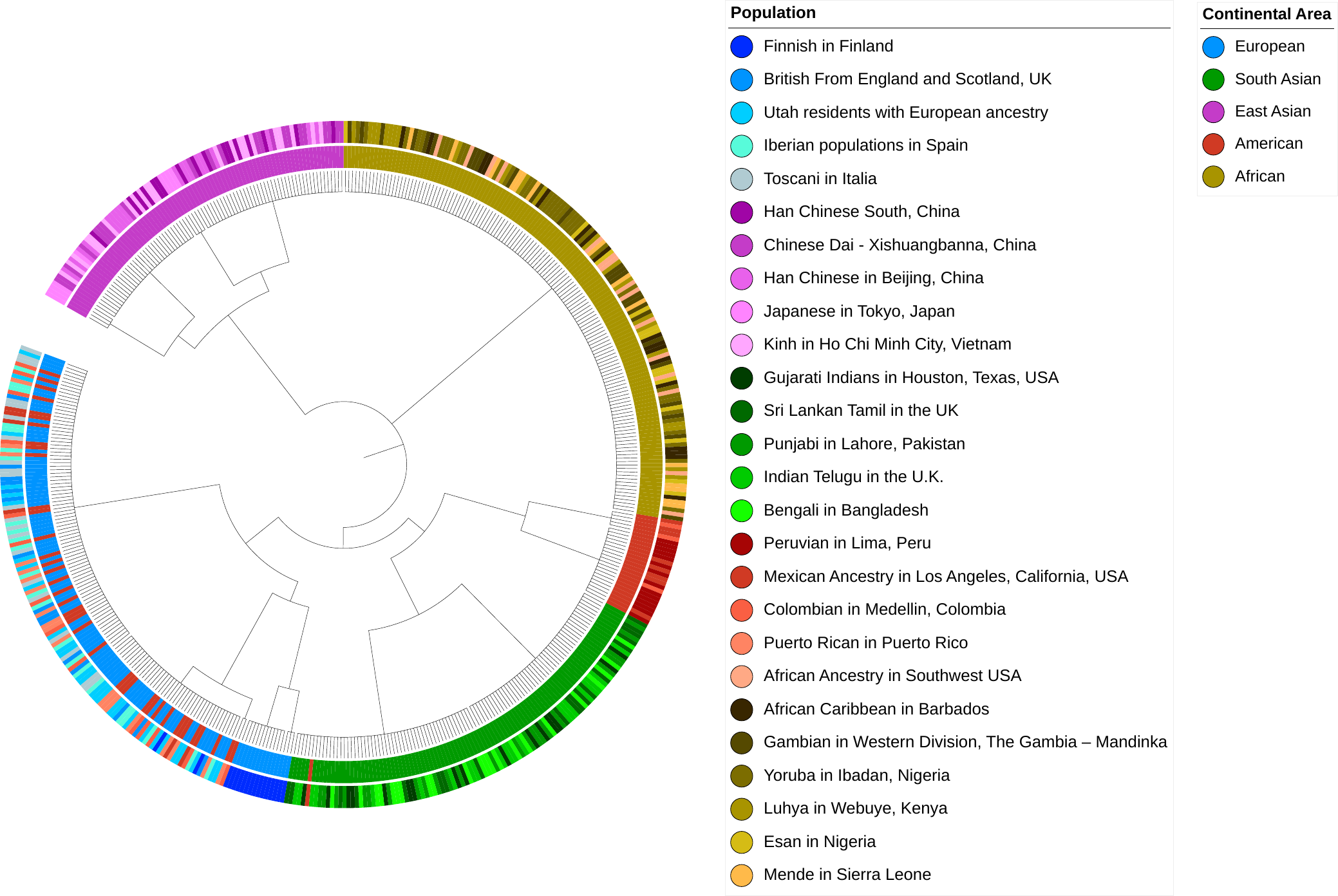}
    \caption{Tree generated from the HGD dataset. The outer coloured markers represent the true labels: continental areas in the inner crown of the circle, and populations in the outer part.}
    \label{fig:tree HGD}
\end{figure*}

In the HGD, each data point is a binary vector corresponding to a human individual, where a 1 is placed in correspondence to the sites that present a mutated gene with respect to a reference sequence and a 0 where the gene is maintained unaltered. For this dataset, we have 2 sets of labels since each individual is associated with a certain continental area (super-population) and population. This two-level characterization of the dataset allows us to test the discriminative power of the RBM at different resolutions. The tree obtained for this dataset is shown in Fig.~\ref{fig:tree HGD}. Overall, the separation of the major labels, corresponding to the continental area, is performed quite well, although sometimes the tree blends together European and American individuals. On the other hand, the substructure corresponding to the different populations is almost never detected by the model at this stage of the training, apart from the isolation of Peruvian and Finnish people. By training an RBM separately on each continental area, we verified that, in some cases, it would be possible to discriminate a subcategory from the others (see, for instance, Fig.~\ref{fig:tree HGD east_asian} of the supplementary text), while in other cases the dataset in itself and the RBM can not resolve the different populations, therefore appearing indistinguishable among each other. We should keep in mind here that, when applying the RBM on populations, we are dealing with a reduced number of samples in the training set and that the impossibility of separating the different populations might be a consequence of the lack of statistical signal. We remark that, since only a sub-part of the genome has been used here, we do not claim at all that these populations would be indistinguishable in general.

\subsection{CPF protein dataset}

In the two cases previously discussed, we had the label information for all the data points. For practical purposes, a more interesting scenario is one in which only a fraction of the data has been annotated. To this end, we considered a biologically relevant dataset consisting of homologous sequences representing the CPF (Cryptochrome/Photolyase) protein. For this dataset, only 86 out of 397 sequences have been experimentally classified, based on their biological function, into 5 different categories \cite{doi:10.1021/cr0204348, PMID:20705454, doi:10.1146/annurev-arplant-042110-103759, jaubert2017light}. Since the amino acid chains are not the same length, the sequences have to be previously aligned in the form of a Multi-Sequence Alignment (MSA) matrix before being fed to the RBM. Each site of the sequences can take one over $N_q = 21$ possible values (20 amino acids + a gap symbol), hence making the dataset particularly suitable to be studied with the Potts-RBM model. Using our tree construction, we may therefore infer a classification for the sequences not previously annotated by looking at the label of the annotated sequences that fall in the same branch.

The algorithm's results on the CPF protein dataset are shown in Fig. \ref{fig:tree FAD}. Coherently with the phylogenetic tree (see Fig. \ref{fig:FAD phylogenetic tree} of the Supplementary Material), the sequences classified as CPD photolyase are found in two distinct main branches of the tree, and a subtree containing photoreceptor sequences is found within the (6-4) photolyase main branch. It is indeed a known fact in biology that the concept of biological function is related, but not identifiable, to the phylogeny of the protein family. In other words, proteins that diverged in their amino acid sequences may still carry the same function. For instance, in Ref.~\cite{10.1093/molbev/msac070} the authors propose a hierarchical classification method, called ProfileView, that is able to correctly identify 11 of the known subgroups of the CPF protein beyond the capabilities of the phylogenetic tree. It is therefore interesting to compare the tree obtained with our method with the phylogenetic one and the tree obtained with ProfileView, for which we remand to the main article~\cite{10.1093/molbev/msac070}. As opposed to the phylogenetic tree, our construction is able to readily identify and separate circadian transcriptional regulators from 6-4 photolyases, although it is not able to separate the animal photoreceptors from the (6-4) photolyases as clearly as it happens with ProfileView. Also, in~\cite{10.1093/molbev/msac070} the photoreceptors are further separated into the plant and plant-like subgroups. The RBM was able to capture the difference between the two categories, although the plant-like sequences are found in two distinct branches of the tree. We notice how the machine very solidly recognises the CRY Pro group, while in the phylogenetic tree it is not clearly distinguishable from the other photolyases. By running different independent trainings using the same hyperparameters, we obtained slightly different tree structures (the training of an RBM is a stochastic process) that nevertheless always conserve the main hierarchical features.

Taking the phylogenetic tree and the ProfileView tree as reference models, we can see that the tree we obtained does not completely separate the different protein functions in the same way, even though it isolates most of the identified functional groups and recovers the phylogeny to some extent. However, we would like to emphasize that our method is able to capture all of these functional signals (and more than just the phylogeny of the sequences) with much less data than ProfileView, which requires the construction of a pool of profile models obtained from the FAD domains of many close sequences throughout the UniProt dataset. Moreover, ProfileView has been specifically optimized for accurate protein function recognition, whereas our method is completely general and has not been optimized for this task.


\begin{figure*}[ht]
    \centering
    \includegraphics[width=\textwidth]{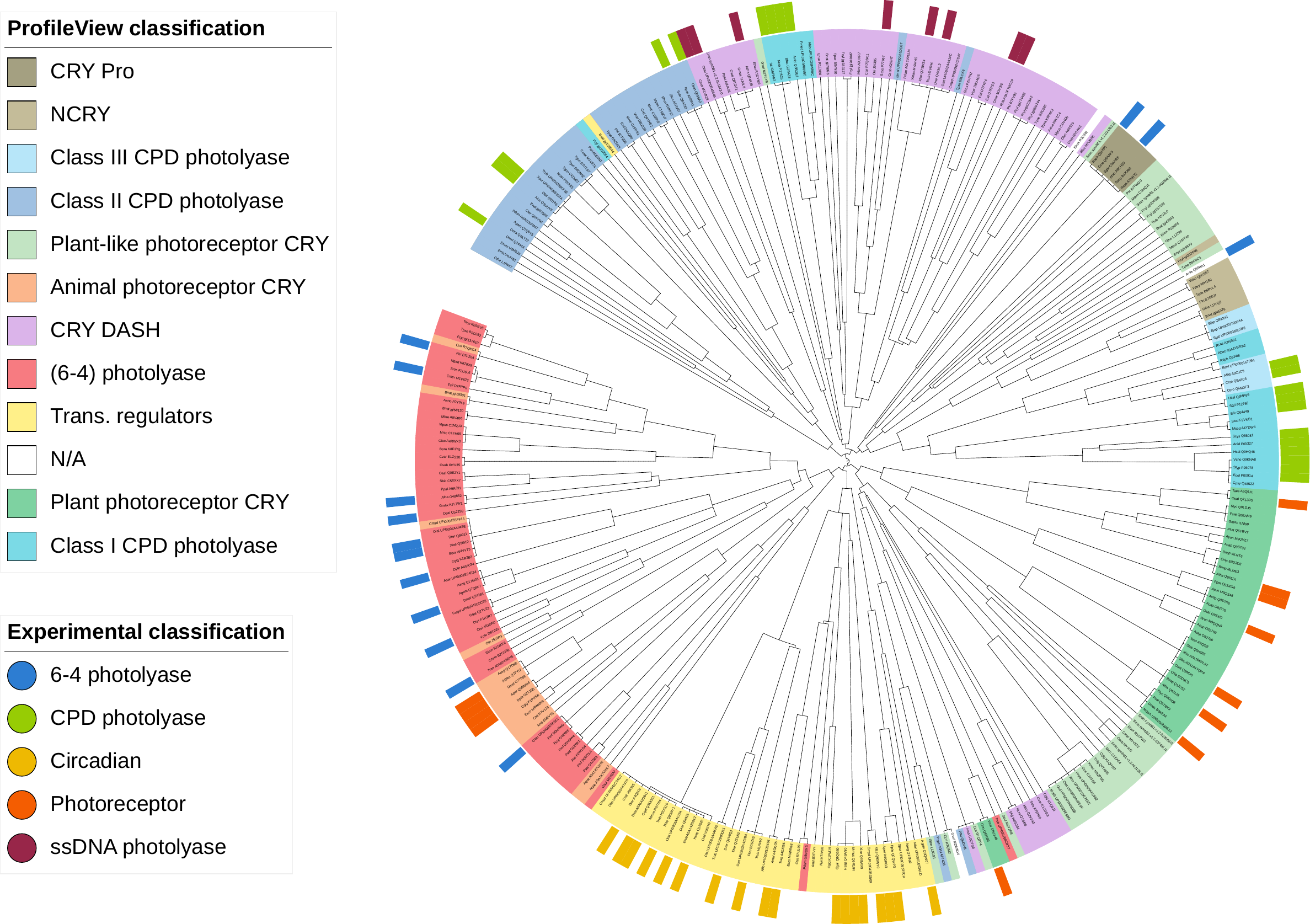}
    \caption{Tree generated from the CPF protein family. The external coloured tags, when available, represent the experimental classification \cite{doi:10.1021/cr0204348, PMID:20705454, doi:10.1146/annurev-arplant-042110-103759, jaubert2017light}, while the colour of the leaves represents the classification into 11 (plus one unidentified) subgroups classified by ProfileView in Ref.~\cite{10.1093/molbev/msac070}.{ Since several ProfileView categories may correspond to the same experimental classification, to avoid confusion, we kept the colours of the two classifications independent to each other.}}
    \label{fig:tree FAD}
\end{figure*}

\section{Discussion}
The presented method scans the evolution of the RBM free energy during training \textit{backwards in time} and uses it to reconstruct the sequential features discovery of the model. The approach we described proved to be far more effective than the na\"ive procedure of clustering the data in the direction of learning. The reason is that the most important splits in the tree correspond to the first branches created by the youngest and ``inexperienced'' models. The feature discrimination capabilities of these models are still quite raw, so many data points that are far from the initial fixed points may be assigned to the wrong feature. Therefore, it is important to start the sequential clustering from the oldest (i.e., ``more experienced'') machines so that the data are gradually pulled through the magnetization space to the correct free energy minima. The sequential clustering of points during tree construction has the effect of not using all the local minima of free energy present at a given age, but only those that find some earlier fixed points in their basin of attraction.

Given that our method is based upon the mean-field approximation, which assumes the model to be in the early stage of the learning, one may allegedly wonder how far we can go with the training and still obtain a reliable approximation of the model's free energy. Since we do not have access to the true free energy of the model, we cannot directly assess the goodness of our second-order approximation. Even so, we can have a clue on the validity of the approximation by comparing the fixed points of the TAP equations with those obtained with the self-consistent equations corresponding to the Gibbs free energy \eqref{eq:Gibbs free energy MT} truncated at the first order in $\beta$, which we call {\em na\"ive} mean-field (nMF). The point of the training at which the two sets of solutions start to diverge can be considered a safe upper bound below which we are guaranteed that our method is working properly. Performing this check for the various datasets reveals that, at the point in which the main features of interest (corresponding to the labelled categories) are separated inside the tree, the fixed points of the first and second-order approximations of the free energy are still very similar. This is particularly true when the dataset is very clusterized. Then, when the two sets of solutions start to change significantly, we observe that the number of nMF fixed points grows much more than that of the TAP solutions, as we can see in Fig. \ref{fig:num fixed points} of the supplementary text.

Another interesting question concerns the variability of the trees obtained with repeated trainings of the same RBM model. Since training the RBM is a stochastic process, we might expect the tree structure to change slightly with different training runs. We verified that there is indeed some degree of variability, as we can see in Fig. \ref{fig:trees comparison} in the supplemental text for the MNIST dataset. However, we also note that some key features of the tree are preserved across experiments, such as the easy separation of digit 0 and the difficult distinction between digits 4 and 9. The same behaviour was observed for other datasets, such as the CPF protein dataset. In this case, we observe that the position of the branches within the tree may vary with repeated experiments, but the hierarchical organization of the subtrees is preserved.

It is worth making a final comment on the possibility of using the training trajectory of the RBM to construct the tree. Since training the RBM can be viewed as an effective annealing of the model, one might want to try to construct the tree by simply using the most trained model and gradually increasing the temperature (i.e., decreasing $\beta$) while iterating the equations from TAP repeatedly until convergence. Although this method gave quite similar results to our original version on the HGD and SED datasets, it proved very inadequate when applied to the MNIST dataset (see Fig. \ref{fig:tree annealing} in the supplementary text). A good reason for using the learning trajectory instead of a trajectory in temperature is that, in the first case, the system undergoes continuous phase transitions~\cite{decelle2017spectral} and the evolution of the probability distribution of the model is smooth. In the second case, on the other hand, the system undergoes first-order phase transitions in temperature, which determine very different free energy landscapes at different temperatures. This could explain why the use of the learning trajectory leads to better results in the hierarchical classification.

{ Finally, one can ask what the result of inference would be if one were to use an RBM trained with a different method than the one we used. For example, one can train the RBM by computing the negative term of the gradient using the TAP equations~\eqref{eq:TAP equations} as proposed in Refs.~\cite{tramel2018deterministic, gabrie2015training}. We have explored this possibility, but we have found empirically that models trained in this way tend to lose the hierarchical properties as the fixed points emerge during learning. We show in Fig.~\ref{fig:tree TAP} of the supplementary text an example of a tree obtained with an RBM trained with the TAP equations. Similarly, trees obtained with non-equilibrium machines, i.e., RBMs trained with non-persistent, non-convergent chains~\cite{decelle2021equilibrium,agoritsas2023explaining}, e.g., with chains initialized with random distributions or the dataset, show some degree of hierarchization in clustering, but are not specific enough to correctly delineate the different families.
}

\section{Conclusions}
In this work, we derived the TAP equations for the Potts RBM defined by the Hamiltonian \eqref{eq:hamiltonian}, and we used this mean-field description of the model to study the evolution of the RBM free energy during training. We then developed a method to infer the hierarchical structures of the data sets from the training trajectory. Using labelled data, we showed that our method is able to generate meaningful trees when applied to four different datasets. In addition, we were able to examine the internal nodes of the tree generated on the SED and MNIST datasets to see how the internal representations of the model change during learning.
Although the tree construction is inherently stochastic, we demonstrated that some relevant features are conserved across different realizations. Finally, we tested our method on a protein dataset and showed that it can provide useful functional information by only capturing the statistics of the input MSA in a purely unsupervised manner, i.e., without any prior knowledge about the data. These results suggest potential promising applications in biology to detect functional signals in homologous protein datasets.

\section*{Acknowledgements}
We would like to thank Giovanni Catania for his help in the early stages of this project in deriving the TAP equations for the Potts RBMs, Alessandra Carbone for useful discussions about the applications of our methods to protein sequences, and Edoardo Sarti for discussions and technical help.



\paragraph{Funding information}
We acknowledge financial support by the Comunidad de Madrid and the Complutense University of Madrid (Spain) through the Atracción de Talento programs (Refs. 2019-T1/TIC-13298 for A.D. and L.R. and 2019-T1/TIC-12776 for B.S.), the Banco Santander and the UCM (grant PR44/21-29937), and Ministerio de Econom\'{\i}a y Competitividad, Agencia Estatal de Investigaci\'on and Fondo Europeo de Desarrollo Regional (FEDER) (Spain and European Union) through the grant PID2021-125506NA-I00.

\bibliography{biblio.bib}

\newpage

\begin{appendix}

\section{Studied Datasets and Technical Details} \label{sec:datasets}
We tested our algorithm on four different datasets:
\begin{itemize}
    \item The MNIST dataset~\cite{lecun1998gradient} consists of grey-scale images representing handwritten digits, from 0 to 9, of dimension $28 \times 28$ pixels. The dataset is thoroughly annotated, namely, we have a label for each sample indicating the represented digit. For our experiments, we extracted 10000 samples from the training set and we applied the tree algorithm to a subset of 500 data samples. The images have been flattened and converted in binary format by setting to 1 the pixels with a normalized value above 0.3 and to 0 otherwise. In this way, the dataset we used was made of 784-dimensional binary vectors.
    \item The Human Genome dataset (HGD)~\cite{10002015global} represents human genetic variations of a population of 5008 individuals sampled from 26 populations in Africa, east Asia, south Asia, Europe and the Americas. Each sample is a sequence of 805 binary variables, $v_i \in \{0, 1\}$, representing the alteration or not of a gene with respect to a reference genetic sequence. For these data, we have a complete two-level hierarchical labelling. At a higher level, the sequences are classified based on the continental origin of the individuals while, at a finer level, we also have a label for the particular population the data was sequenced from. This allows us to test the discriminative capabilities of the RBM at different resolutions. We trained the RBM on 4508 samples.
    \item The Synthetic Evolutionary Data (SED) dataset has been artificially created by us with the aim of simulating a controlled evolutionary process fully driven by random mutations. We started with an ancestor sequence of length $N_v=805$ and we produced 5 heirs by flipping at least one site of the sequence. Then, we associated a label with each one of the heirs, and we iterated the following evolutionary process: given a sequence, we generated a random number between 1 and 5 which corresponds to the number of children. Each child is obtained by flipping each site of the mother sequence with probability $p=1/N_v$, and it is added to the dataset if the sequence was not already present. Each child inherits the label of the mother sequence. We repeated this process at each generation until we obtained a dataset of 4508 sequences.
    \item Finally, we tested the algorithm on a dataset of homologous protein sequences. The dataset we consider is made of 397 amino acid sequences representing the Cryptochrome/Photolyase (CPF) protein. Proteins belonging to the CPF family share the same fold, yet perform several different functions. Cryptochromes are divided into photoreceptors that activate specific pathways in presence of light and light-independent circadian clock transcriptional regulators, while photolyases are light-activated enzymes encharged with repairing UV-damaged DNA. In particular, 86 of the available sequences have been experimentally classified into 5 categories according to their biological function \cite{doi:10.1021/cr0204348, PMID:20705454, doi:10.1146/annurev-arplant-042110-103759, jaubert2017light}: circadian, photoreceptor, (6-4) photolyase, CPD photolyase and ssDNA photolyase. Finally, we validated our model by also considering the finer classification in 11 categories determined in \cite{le2008representational}.
    
    To be processed by the RBM, the sequences must be previously aligned in a matrix called Multi-Sequence Alignment (MSA). The alignment is performed by inserting a special symbol, the \textit{gap} `-', a proper number of times within adjacent sites of the sequences. Each site can therefore assume one over 21 possible values (20 amino acids + the gap). After the alignment, the length of the obtained sequences was $N_v = 7403$. To perform the MSA we used the software \href{https://mafft.cbrc.jp/alignment/software/}{MAFFT} with a large \textit{ep} parameter of 1 to penalize long alignments.

    Concerning the weights used for computing the weighted average in Eq.~\ref{eq:LL weighted} in this dataset, we first had to define a set of clusters by gathering all sequences with sequence identity greater than 0.8, and then we applied the same criterion described above. The clustering has been done using the software \href{https://github.com/soedinglab/MMseqs2}{MMseqs2}.

    All the trees shown in the article have been generated using the \href{https://itol.embl.de/}{iTOL} software.
\end{itemize}

\subsection{Technical details of the RBM models used}
In Tab.~\ref{tab:RBMs specifics} we report the hyper-parameters of the RBMs used for generating the trees reported in the main article.

\begin{table*}[ht]
    \centering
    \begin{tabular}{cccccccc}
    \toprule
    \textbf{tree} & \textbf{dataset} & \textbf{epochs} & \thead{\textbf{minibatch} \\ \textbf{size}} & \thead{\textbf{total gradient} \\ \textbf{updates}} & $\mathbf{N_{\mathrm{Gibbs}}}$ & \thead{\textbf{learning} \\ \textbf{rate}} & $\mathbf{N_h}$  \\
    \midrule
    Fig.~\ref{fig:collage}-F & SED & 10000 & 512 & $9 \cdot 10^4$ & 100 & $10^{-2}$ & 512 \\
    \midrule
    Figs.~\ref{fig:tree images MNIST}, \ref{fig:tree MNIST} & MNIST & 10000 & 512 & $2 \cdot 10^5$ & 100 & $10^{-3}$ & 512\\
    \midrule
    Fig.~\ref{fig:tree HGD} & HGD & 10000 & 512 & $9 \cdot 10^4$ & 100 & $10^{-3}$ & 512 \\
    \midrule
    Fig.~\ref{fig:tree FAD} & CPF & 10000 & 128 & $3 \cdot 10^4$ & 100 & $10^{-3}$ & 512 \\
    \bottomrule
    \end{tabular}
    \caption{Technical specifics of the RBMs used for generating the trees reported in the main text. All machines have been trained in PCD mode~\cite{decelle2021equilibrium}. Only for the tree of Fig.~\ref{fig:collage}-F, a binary RBM has been used in place of the Potts version described in the main text.}
    \label{tab:RBMs specifics}
\end{table*}

{
\subsection{Projecting the dataset along the weight matrix principal directions}\label{ap:proj}
In order to project a dataset made of categorical variables along the first principal directions of the weight matrix, we first have to transpose the tensor $\bm w$ along the first 2 dimensions and then reshape it from a $N_v \times N_q \times N_h$ tensor to a $(N_q \cdot N_v) \times N_h$ matrix:
\begin{align}
    \bm w \in  \mathbb{R}^{N_v \times N_q \times N_h} & \xrightarrow{\text{transpose}} \hat{\bm w} \in \mathbb{R}^{N_q \times N_v \times N_h}\nonumber \\ 
    & \xrightarrow{\text{reshape}} \tilde{\bm w} \in \mathbb{R}^{(N_q \cdot N_v) \times N_h}. \nonumber
\end{align}

Then, we consider the Singular Value Decomposition (SVD) of $\tilde{\bm w}$,
\begin{equation}
    \tilde{\bm w} = U S V^T,
\end{equation}
where $U \in \mathbb{R}^{(N_q \cdot N_v) \times (N_q \cdot N_v)}$, $S$ is a $(N_q \cdot N_v) \times N_h$ matrix with the singular values of $\bm w$ in the diagonal and $V \in \mathbb{R}^{N_h \times N_h}$. The next step is to compute the one-hot representation of the dataset, which consists in associating to each categorical variable $q$ a $N_q$-dimensional vector that has value 1 in position $q$ and 0 otherwise:
\begin{align}
    \mathcal{D} \in  \mathbb{R}^{M \times N_v} & \xrightarrow{\text{one-hot}} \hat{\mathcal{D}} \in \mathbb{R}^{M \times N_v \times N_q}\nonumber \\ 
    & \xrightarrow{\text{reshape}} \tilde{\mathcal{D}} \in \mathbb{R}^{M \times (N_v \cdot N_q)}. \nonumber
\end{align}

Finally, the projected dataset is given by
\begin{equation}
    \mathcal{D}_{\mathrm{proj}} = \tilde{\mathrm{D}} U \in \mathbb{R}^{M \times (N_v \cdot N_q)},
\end{equation}
where each row is a vector associated to a data sample whose components correspond to the projections of the one-hot representation of the data point along the principal directions of the weight matrix.
}

\section{TAP equations for the Potts RBM}
\label{app:TAP equations for the Potts RBM}
We start by considering a Hamiltonian of the kind:
\begin{equation}
    -\beta E_\beta(\pmb{v}, \pmb{h}; \pmb{\theta}) = \sum_{i q} a_i^q \delta_{v_i, q} + \sum_\mu b_\mu h_{\mu} +\beta \sum_{i q \mu}\delta_{v_i, q}w_{i \mu }^q h_\mu,
\end{equation}
where the inverse-temperature parameter, $\beta$, regulates the interactions among the variables. Our goal is to study the metastable minima of the free energy in the high-temperature regime $\beta \rightarrow 0$, which is equivalent to considering the limit in which the eigenvalues of the weight matrix are vanishing.

Let us introduce two sets of external fields, $\{\phi_j^p\}$ and $\{\psi_\mu\}$, which act respectively on the visible and hidden variables and depend on the inverse-temperature $\beta$. We denote the total external field vector as $\pmb{B} = \{B_i\} = (\{\phi_j^p\},\{\psi_\mu\})$, with $i \in \{1, \dots, N_v \cdot N_q + N_h\}$. The free energy of this extended system is
\begin{align}
    - \beta F_{\beta}(\pmb{B}) =&\log \sum_{\pmb{v}, \pmb{h}} \exp\biggl[-\beta E_\beta(\pmb{v}, \pmb{h}; \pmb{\theta}) + \sum_{i q} \phi_i^q(\beta) \delta_{v_i, q} + \nonumber\\ &+ \sum_\mu \psi_\mu(\beta) h_\mu\biggr],
    \label{eq:free energy extended system}
\end{align}
where we omitted the free energy's dependence from the model's parameters $\pmb{\theta}$. The local magnetizations in the presence of these fields, $\pmb{M} = \{M_i\} = (\{f_j^p\},\{m_\mu\})$, are
\begin{eqnarray}
    &&f_i^q = \pa{\delta_{v_i, q}}_{\pmb B} = -\beta \frac{\partial}{\partial \phi_i^q} F_\beta(\pmb{B}) \quad \text{and} \nonumber\\ \quad &&m_\mu = \pa{h_\mu}_{\pmb B} = -\beta \frac{\partial}{\partial \psi_\mu} F_\beta(\pmb{B}).
    \label{eq:magnetizations definition}
\end{eqnarray}
Notice that the free energy in equation \eqref{eq:free energy extended system} is a concave function of the external fields. To see this, we can represent a state of the system with the vector $\pmb{x} = (\pmb{\delta}^{\mathrm{oh}}, \pmb{h}) \in \mathbb{R}^{N_v \cdot N_q + N_h}$, where $\pmb{\delta}^{\mathrm{oh}} \in \mathbb{R}^{N_v \cdot N_q}$ is the one-hot representation of the Potts states. Then, the Hessian of the free energy with respect to the fields is
\begin{align}
    \frac{\partial^2 F_\beta(\pmb B)}{\partial B_j \partial B_i} &= -\frac{1}{\beta} \frac{\partial M_i}{\partial B_j} = - \frac{1}{\beta}\pa{(x_i - M_i)(x_j - M_j)}_{\pmb{B}} \nonumber\\ \text{for} \quad &i, j \in \{1, \dots, N_v \cdot N_q + N_h\},
\end{align}
which is semi-negative defined.

We can describe the system in terms of the magnetizations by considering the Gibbs free energy, which is obtained by taking the Legendre transform of \eqref{eq:free energy extended system}:
\begin{widetext}
\begin{align}
    -\beta G_\beta(\pmb{M}) &= -\beta \max_{\pmb{B}} \biggl[F_\beta(\pmb{B}) + \frac{1}{\beta} \sum_{i q} \phi_i^q(\beta) f_i^q +\frac{1}{\beta} \sum_\mu \psi_\mu(\beta) m_\mu\biggr] = \nonumber \\
    &= \min_{\pmb{B}} \biggl[\log \sum_{\pmb{v}, \pmb{h}} \exp \biggl(\sum_{i q} a_i^q \delta_{v_i, q} + \sum_\mu b_\mu h_{\mu} +\beta \sum_{i q \mu}\delta_{v_i, q}w_{i \mu }^q h_\mu + \sum_{i q} \phi_i^q(\beta)(\delta_{v_i, q} - f_i^q) + \sum_\mu \psi_\mu(\beta)(h_\mu - m_\mu)\biggr)\biggr];
\end{align}
\end{widetext}
in this way the fields $\pmb{B} = \pmb{B}(\pmb{M})$ satisfy equations \eqref{eq:magnetizations definition}, namely they enforce the magnetizations $\pmb{M}$.
The solution $\pmb{B}(\pmb{M})$ must satisfy
\begin{widetext}
\begin{equation}
    B_i = \beta \frac{\partial}{\partial M_i} G_\beta(\pmb{M}), \qquad \frac{\partial^2 G_\beta(\pmb{M})}{\partial M_j \partial M_i} = \frac{1}{\beta} \frac{\partial}{\partial M_j} B_i(\pmb{M}) = \frac{1}{\beta}\pt{\frac{\partial M_j}{\partial B_i}}^{-1} = - \pt{\frac{\partial^2 F_\beta(\pmb{B})}{\partial B_i \partial B_j}}^{-1},
    \label{eq:Gibbs free energy is convex}
\end{equation}
\end{widetext}
meaning that the Gibbs free energy is a convex function of the magnetizations. The free energy is the inverse Legendre transform of $G_\beta(\pmb{M})$,
\begin{align}
    -\beta F_\beta(\pmb{B}) = -\beta \min_{\pmb{M}} \biggl[G_\beta(\pmb{M}) &- \frac{1}{\beta} \sum_{i q} \phi_i^q(\beta) f_i^q + \nonumber\\ &- \frac{1}{\beta} \sum_\mu \psi_\mu(\beta) m_\mu\biggr],
    \label{eq:inverse Legendre transform}
\end{align}
and the solution $\pmb{M}(\pmb{B})$ satisfies equation \eqref{eq:Gibbs free energy is convex} and it is the global minimum of the Gibbs free energy. From equation \eqref{eq:inverse Legendre transform} we can see that, by turning off the external fields $\pmb{B}$, we retrieve the free energy of the original system at the minimum of the Gibbs free energy:
\begin{equation}
    A_\beta = -\frac{1}{\beta} \sum_{\pmb{v}, \pmb{h}} e^{-\beta E_\beta(\pmb{v}, \pmb{h}, \pmb{\theta})} = F_\beta(\pmb{B} = \pmb{0}) = \min_{\pmb{M}} \pq{G_\beta(\pmb{M})}.
    \label{eq:free energy stationary conditions}
\end{equation}

At this point, we have two problems. The first problem is that the Gibbs free energy is a convex function of the magnetizations, meaning that we can not use it directly to identify the metastable minima of the free energy. Moreover, we are prevented from computing it in the first place, because it would be as difficult to evaluate as the original free energy. The two issues can be overcome if we restrict ourselves to the high-temperature limit, corresponding to the early stage of the learning when the RBM has not yet learnt the modes of the dataset. In this regime, we can approximate the Gibbs free energy with a Taylor expansion around $\beta = 0$. For ease of notation, let us define the Gibbs free entropy as $\Gamma_\beta(\pmb{M}) = -\beta G_\beta(\pmb{M})$. We want to compute the second-order expansion:
\begin{equation}
    \Gamma_\beta^{(2)}(\pmb{M}) = \Gamma_0(\pmb{M}) + \beta \evalat{\frac{\partial \Gamma_\beta(\pmb{M})}{\partial \beta }}{\beta=0} + \frac{\beta^2}{2} \evalat{\frac{\partial^2 \Gamma_\beta(\pmb{M})}{\partial \beta^2}}{\beta=0}.
    \label{eq:expansion of the Gibbs free entropy}
\end{equation}
The high-temperature approximation has the effect of breaking the degeneracy of the Gibbs free energy, hence allowing us to detect the metastable pure states. Moreover, the expansion \eqref{eq:expansion of the Gibbs free entropy} can be directly computed, because at $\beta=0$ the interactions among the variables disappear, and the probability measure factorizes.
Given an observable $\mathcal{O}$, we can define the high-temperature average as
\begin{widetext}
\begin{equation}
    \pa{\mathcal{O}} = \frac{\sum_{\pmb{v}, \pmb{h}} \mathcal{O}(\pmb{v}, \pmb{h}) \exp \biggl(\sum_{i q} (a_i^q + \phi_i^q(0)) \delta_{v_i, q} + \sum_\mu (b_\mu + \psi_\mu(0)) h_{\mu} \biggr)}{\sum_{\pmb{v}, \pmb{h}} \exp \biggl(\sum_{i q} (a_i^q + \phi_i^q(0)) \delta_{v_i, q} + \sum_\mu (b_\mu + \psi_\mu(0)) h_{\mu} \biggr)}.
\end{equation}
\end{widetext}
In particular, we can compute the magnetizations in the high-temperature regime as
\begin{align}
    m_\mu &= \pa{h_\mu} = \frac{\sum_{h_\mu = 0, 1} h_\mu \exp \pt{\sum_\mu h_\mu (\psi_\mu(0) + b_\mu)}}{\sum_{h_\mu = 0, 1} \exp \pt{\sum_\mu h_\mu (\psi_\mu(0) + b_\mu)}}= \nonumber\\ &= \frac{1}{1 + e^{-b_\mu - \psi_\mu(0)}} = \mathrm{sigmoid}(b_\mu + \psi_\mu(0)) \label{eq:sigmoid m},
\end{align}
\begin{align}
    f_i^q &= \pa{\delta_{v_i, q}} = \frac{\sum_{v} \delta_{v, q} \exp \pt{\sum_{p} \delta_{v, p} (\phi_i^{p}(0) + a_i^{p})}}{\sum_{v} \exp \pt{\sum_{p} \delta_{v, p} (\phi_i^{p}(0) + a_i^{p})}}= \nonumber\\ & = \frac{e^{a_i^q + \phi_i^q(0)}}{\sum_{p} e^{a_i^p + \phi_i^{p}(0)}} = \mathrm{softmax}(a_i^q + \phi_i^q(0))\label{eq:softmax f}.
\end{align}
We now proceed by computing the various terms of equation \eqref{eq:expansion of the Gibbs free entropy}. For ease of notation, let us define
\begin{gather}
    \mathcal{I} = \sum_{i q \mu}\delta_{v_i, q}w_{i \mu }^q h_\mu \\
    \mathcal{F}(\beta) = \sum_{i q} \phi_i^q(\beta)(\delta_{v_i, q} - f_i^q) + \sum_\mu \psi_\mu(\beta)(h_\mu - m_\mu),
\end{gather}
so that the Gibbs free entropy can be written as
\begin{align}
    \Gamma_\beta(\pmb{M}) = \min_{\pmb{B}} \biggl[\log \sum_{\pmb{v}, \pmb{h}} \exp \biggl(\beta \mathcal{I} +& \mathcal{F}(\beta) + \sum_{i q} a_i^q \delta_{v_i, q} + \nonumber\\ + &\sum_\mu b_\mu h_{\mu}\biggl)\biggl].
\end{align}

\paragraph{Entropic term}
The zero-order term corresponds to the mean-field entropy of the model, and it is given by:

\begin{align}
    \Gamma_0(\pmb{M}) =& -\sum_{i q} \phi_i^q(0) f_i^q - \sum_{\mu} \psi_\mu(0) m_\mu +  \nonumber\\ & + \sum_i \log \pq{\sum_{v} \exp \pt{a_i^v + \phi_i^{v}(0)}} +  \nonumber\\ & + \sum_\mu \log \pq{1 + \exp\pt{b_\mu + \psi_\mu(0)}}.
    \label{eq:Gamma0}
\end{align}

In order to make explicit the dependence on the magnetizations $\{m_\mu\}$ and $\{f_i^q\}$ into equation \eqref{eq:Gamma0}, we need to invert equations \eqref{eq:sigmoid m} and \eqref{eq:softmax f}. The first equation leads straightforwardly to
\begin{equation}
    \psi_\mu(0) = \log \pt{\frac{m_\mu}{1 - m_\mu}} - b_\mu,
    \label{eq:inv_sigmoid m}
\end{equation}
while from the second equation we get
\begin{equation}
    \phi_i^q(0) = \log f_i^q -a_i^q + c_i
\end{equation}
where $c_i = \log \pq{\sum_p \exp\pt{a_i^p + \phi_i^p(0)}}$. This last arbitrary constant comes from the fact that the Potts model displays a gauge invariance~\cite{cocco2018inverse}. A convenient choice to fix it is the \textit{lattice-gas gauge}, for which we impose
\begin{equation}
    a_i^{N_q} = \phi_i^{N_q} = w_{i \mu}^{N_q} = 0.
\end{equation}
In this way, we obtain the inverse relation of the softmax function \eqref{eq:softmax f}:
\begin{equation}
    \phi_i^q(0) = \log \pt{\frac{f_i^q}{f_i^{N_q}}} - a_i^q.
\end{equation}

By inserting these relations into equation \eqref{eq:Gamma0} and remembering the normalization condition $\sum_q f_i^q = 1$, we obtain the familiar expression of the mean-field Shannon entropy of a set of binary and categorical variables plus two interaction terms between the magnetizations and the local fields of the model,
\begin{align}
    \Gamma_0(\pmb{M}) =& \sum_{i q}f_i^q a_i^q + \sum_\mu m_\mu b_\mu - \sum_{i q} f_i^q \log f_i^q + \nonumber\\ &- \sum_\mu \pq{m_\mu \log m_\mu + (1 - m_\mu) \log (1-m_\mu)}.
    \label{eq:Gamma0 explicit}
\end{align}

\paragraph{First order term}
The first derivative is readily obtained as:
\begin{align}
    \evalat{\frac{\partial \Gamma_\beta}{\partial \beta}}{\beta=0} = \pa{\mathcal{I}} + \underbrace{\pa{\partial_\beta \mathcal{F}(0)}}_{=0} = \sum_{i q \mu} f_i^q w_{i \mu}^q m_\mu 
\end{align}
 which is nothing but the na\"ive mean field interaction energy.
 
 \paragraph{Second order term}
Given an observable $\mathcal{O}(\beta)$, the derivative of its average at inverse-temperature $\beta$ can be computed as
\begin{align}
    \frac{\partial}{\partial \beta} \pa{\mathcal{O}(\beta)}_\beta  =&\pa{\partial_\beta \mathcal{O}(\beta)}_\beta + \pa{\mathcal{O}(\beta) \mathcal{I}}_\beta + \pa{\mathcal{O}(\beta)\partial_\beta \mathcal{F}(\beta)}_\beta + \nonumber\\& - \pa{\mathcal{O}(\beta)}_\beta \pa{\mathcal{I}}_\beta - \pa{\mathcal{O}(\beta)}\pa{\partial_\beta \mathcal{F}(\beta)}_\beta.
\end{align}
For $\beta=0$ the expression simplifies into:
\begin{align}
    \frac{\partial}{\partial \beta} \pa{\mathcal{O}(0)}  =&\pa{\partial_\beta \mathcal{O}(0)} + \pa{\mathcal{O}(0) \mathcal{I}} + \pa{\mathcal{O}(0)\partial_\beta \mathcal{F}(0)}+ \nonumber\\ &- \pa{\mathcal{O}(0)}_\beta \pa{\mathcal{I}},
\end{align}
so that the second derivative of the Gibbs free entropy is
\begin{align}
    \evalat{\frac{\partial^2 \Gamma_\beta}{\partial \beta^2}}{\beta=0} &= \frac{\partial}{\partial \beta} \pa{\mathcal{I}} = \pa{\mathcal{I}^2} - \pa{\mathcal{I}}^2 + \pa{\mathcal{I} ~\partial_\beta \mathcal{F}(0)}.
    \label{eq:d2Gamma}
\end{align}

To proceed further with the computation, we need to use the following pair correlation functions computed at $\beta=0$:
\begin{align}
\begin{split}
    \pa{h_\mu h_\nu} &= m_\mu \delta_{\mu \nu} + (1 - \delta_{\mu \nu}) m_\mu m_\nu,\\
    \pa{\delta_{v_i, q} \delta_{v_j, p}} &= f_i^q \delta_{ij} \delta_{qp} + (1 - \delta_{ij}) f_i^q f_j^p.
\end{split}
\end{align}
Also, we need to compute the first derivative of the external fields with respect to $\beta$ at $\beta = 0$. These can be easily obtained using the Schwarz theorem:
\begin{align}
    \begin{split}
    \partial_\beta \phi_i^q(0) &= - \evalat{\frac{\partial^2 \Gamma_\beta}{\partial \beta \partial f_i^q}}{\beta=0} = - {\evalat{\frac{\partial^2 \Gamma_\beta}{\partial f_i^q \partial \beta}}{\beta=0}} = - \sum_\mu w_{i \mu}^q m_\mu,\\
    \partial_\beta \psi_\mu(0) &= - {\evalat{\frac{\partial^2 \Gamma_\beta}{\partial \beta \partial m_\mu}}{\beta = 0}} = - {\evalat{\frac{\partial^2 \Gamma_\beta}{\partial m_\mu \partial \beta}}{\beta = 0}} = - \sum_{i q} f_i^q w_{i \mu}^q.
    \end{split}
\end{align}

By expanding the expression \eqref{eq:d2Gamma}, after many simplifications we end up with the following second-order Taylor expansion of the Gibbs free entropy:
\begin{widetext}
\begin{alignat}{3}
    \Gamma_\beta^{(2)}(\pmb{M}) &= &&\Gamma_0(\pmb{M}) + \beta \evalat{\frac{\partial \Gamma_\beta(\pmb{M})}{\partial \beta }}{\beta=0} + \frac{\beta^2}{2} \evalat{\frac{\partial^2 \Gamma_\beta(\pmb{M})}{\partial \beta^2}}{\beta=0} = \nonumber \\
    &= && \sum_{i q} f_i^q a_i^q + \sum_\mu m_\mu b_\mu - \sum_{i q} f_i^q \log f_i^q - \sum_\mu \pq{m_\mu \log m_\mu + (1 - m_\mu) \log (1-m_\mu)} + \nonumber \\
    & &&+ \beta \sum_{i q \mu} f_i^q w_{i \mu}^q m_\mu + \nonumber\\ 
    & &&+ \frac{\beta^2}{2} \sum_\mu (m_\mu - m_\mu^2) \pq{\sum_{i q} (w_{i \mu}^q)^2  f_i^q - \sum_{i} \pt{\sum_q w_{i \mu}^q f_i^q}^2}.
    \label{eq:TAP Gibbs free energy}
\end{alignat}
\end{widetext}

The previous expression is a valid approximation of the true free energy of the model only when the magnetizations fulfil the stationary conditions \eqref{eq:free energy stationary conditions}:
\begin{equation}
    \frac{\partial \Gamma_\beta^{(2)}}{\partial m_\mu} = 0, \qquad \frac{\partial \Gamma_\beta^{(2)}}{\partial f_i^q} = 0.
    \label{eq:stat_conds}
\end{equation}
When computing the stationary condition with respect to the visible magnetization, one has to take care of the fact that the normalization condition on $\{f_i^q\}$ leaves us with only $N_q-1$ independent colours, so that $f_i^{N_q} = 1 - \sum_{q=1}^{N_q-1} f_i^q$. This has to be considered only in the entropic term, since elsewhere the magnetization couples with the weight matrix that is zero along the colour $N_q$ because of the lattice-gas gauge.

The full computation of the stationary conditions \eqref{eq:stat_conds} at $\beta=1$ yields
\begin{widetext}
\begin{align}
\begin{split}
    \log \pt{\frac{m_\mu}{1 - m_\mu}} - b_\mu &= \sum_{i q} f_i^q w_{i \mu}^q - \pt{m_\mu - \frac{1}{2}} \pq{\sum_{i q} (w_{i \mu}^q)^2  f_i^q - \sum_{i} \pt{\sum_q w_{i \mu}^q f_i^q}^2},\\
    \log \pt{\frac{f_i^q}{f_i^{N_q}}} - a_i^q &= \sum_\mu m_\mu w_{i \mu}^q + \frac{1}{2}\sum_\mu (m_\mu - m_\mu^2) \pq{(w_{i \mu}^q)^2  - 2 w_{i \mu}^q \sum_{p}f_i^p w_{i \mu}^p},
\end{split}
\end{align}
\end{widetext}
where at the left-hand side we recognize respectively the inverse functions of the sigmoid \eqref{eq:sigmoid m} and of the softmax \eqref{eq:softmax f}. Eventually, we end up with the following self-consistent TAP equations:

\begin{widetext}
\begin{align}
\begin{split}
    m_\mu &= \mathrm{sigmoid}\pg{b_\mu + \sum_{i q} f_i^q w_{i \mu}^q - \pt{m_\mu - \frac{1}{2}} \pq{\sum_{i q} (w_{i \mu}^q)^2  f_i^q - \sum_{i} \pt{\sum_q w_{i \mu}^q f_i^q}^2}},\\
    f_i^q &= \mathrm{softmax}\pg{a_i^q + \sum_\mu m_\mu w_{i \mu}^q + \frac{1}{2}\sum_\mu (m_\mu - m_\mu^2) \pq{(w_{i \mu}^q)^2  - 2 w_{i \mu}^q \sum_{p}f_i^p w_{i \mu}^p}}.
\end{split}
\end{align}
\end{widetext}


\section{Supplementary figures}
\label{app:supplementary figures}
\begin{figure*}[ht]

    \centering
    \includegraphics[width=\textwidth]{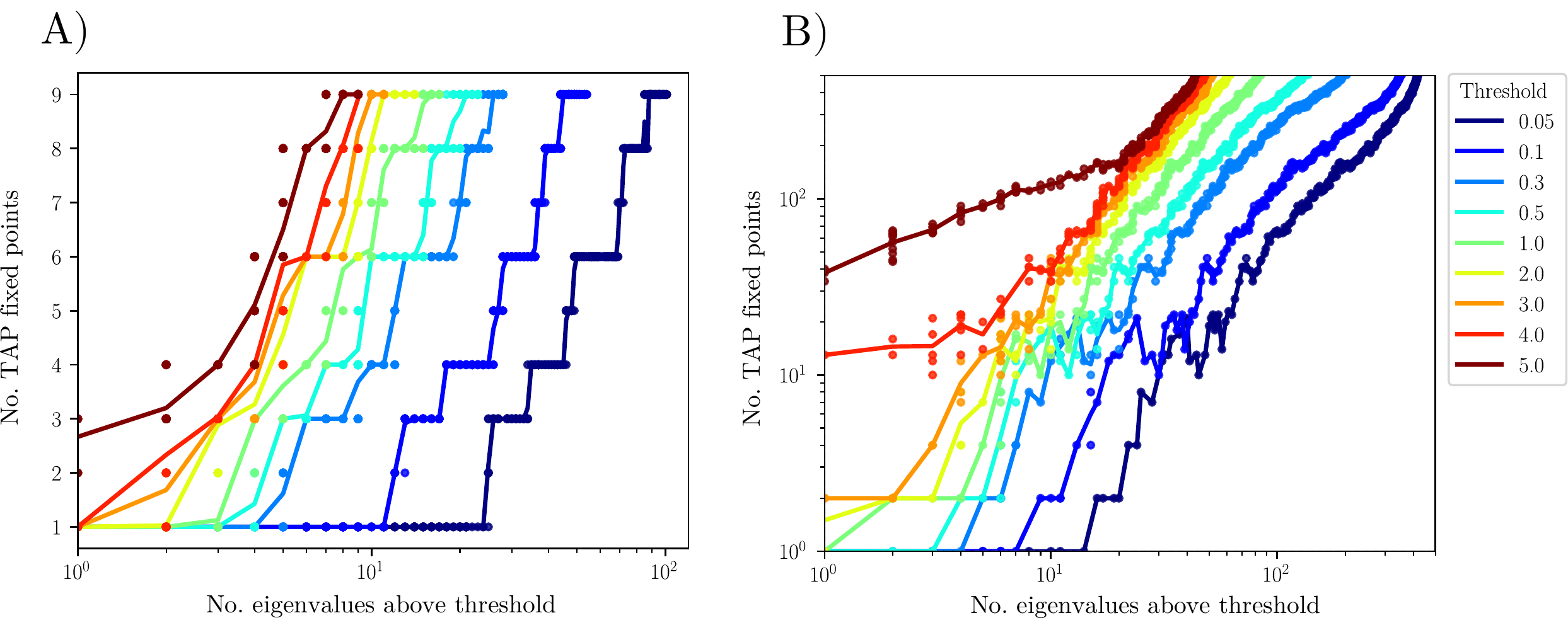}
    \caption{For each model saved during the training, we recorded the number of TAP fixed points and the number of eigenvalues of the weight matrix above different thresholds. The different colours correspond to different thresholds, and the continuous lines represent the average number of TAP fixed points expected when a given number of eigenvalues are above the threshold. Panel A) shows the results obtained on the SED dataset, whereas panel B) shows the results on the MNIST dataset.}
    \label{fig:eig vs TAP}
\end{figure*}

\begin{figure*}[ht]
    \centering
    \includegraphics[width=\textwidth]{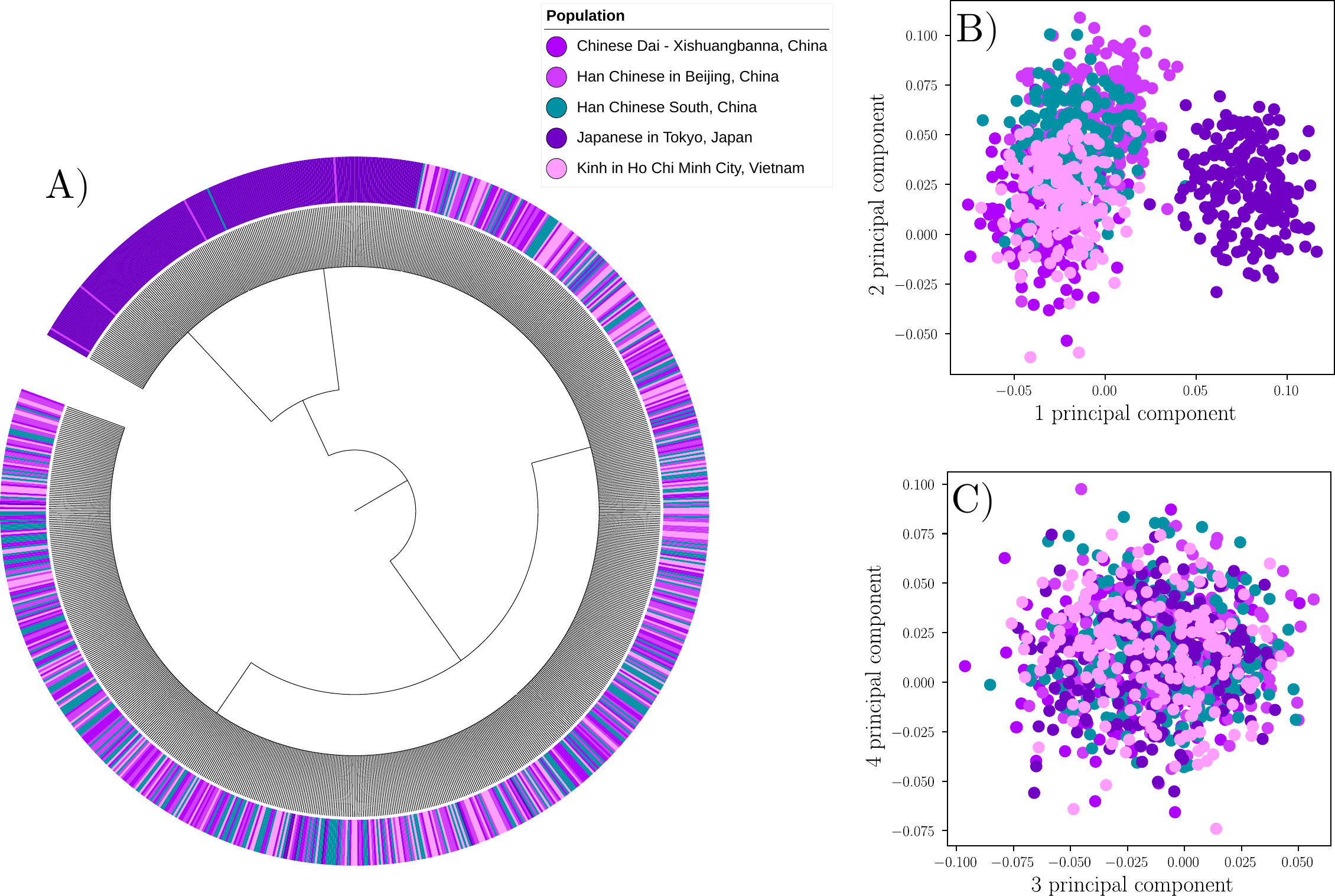}
    \caption{A) Tree generated on all the HGD data denoted as ``East asian''. The tree is able to only distinguish the data labelled as ``Japanese in Tokyo, Japan'' from the others. B) and C) Projection of the dataset along the first and second principal components (panel B) and along the third and fourth principal components (panel C).}
    \label{fig:tree HGD east_asian}
\end{figure*}

\begin{figure*}[ht]
    \centering
    \includegraphics[width=\textwidth]{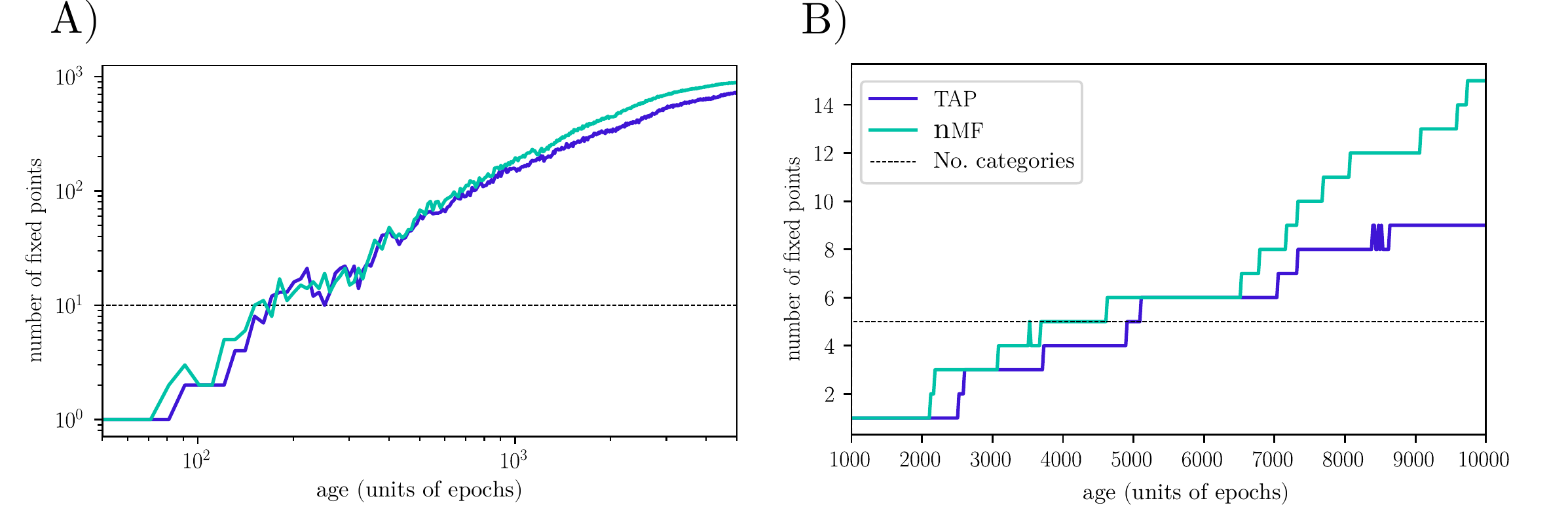}
    \caption{Number of fixed points of the approximated free energy on the MNIST dataset (panel A) and on the SED dataset (panel B) as a function of the RBM's age. Notice that plot A is in log-log scale, whereas plot B is in linear-linear scale.}
    \label{fig:num fixed points}
\end{figure*}

\begin{figure*}[ht]
    \centering
    \includegraphics[width=\textwidth]{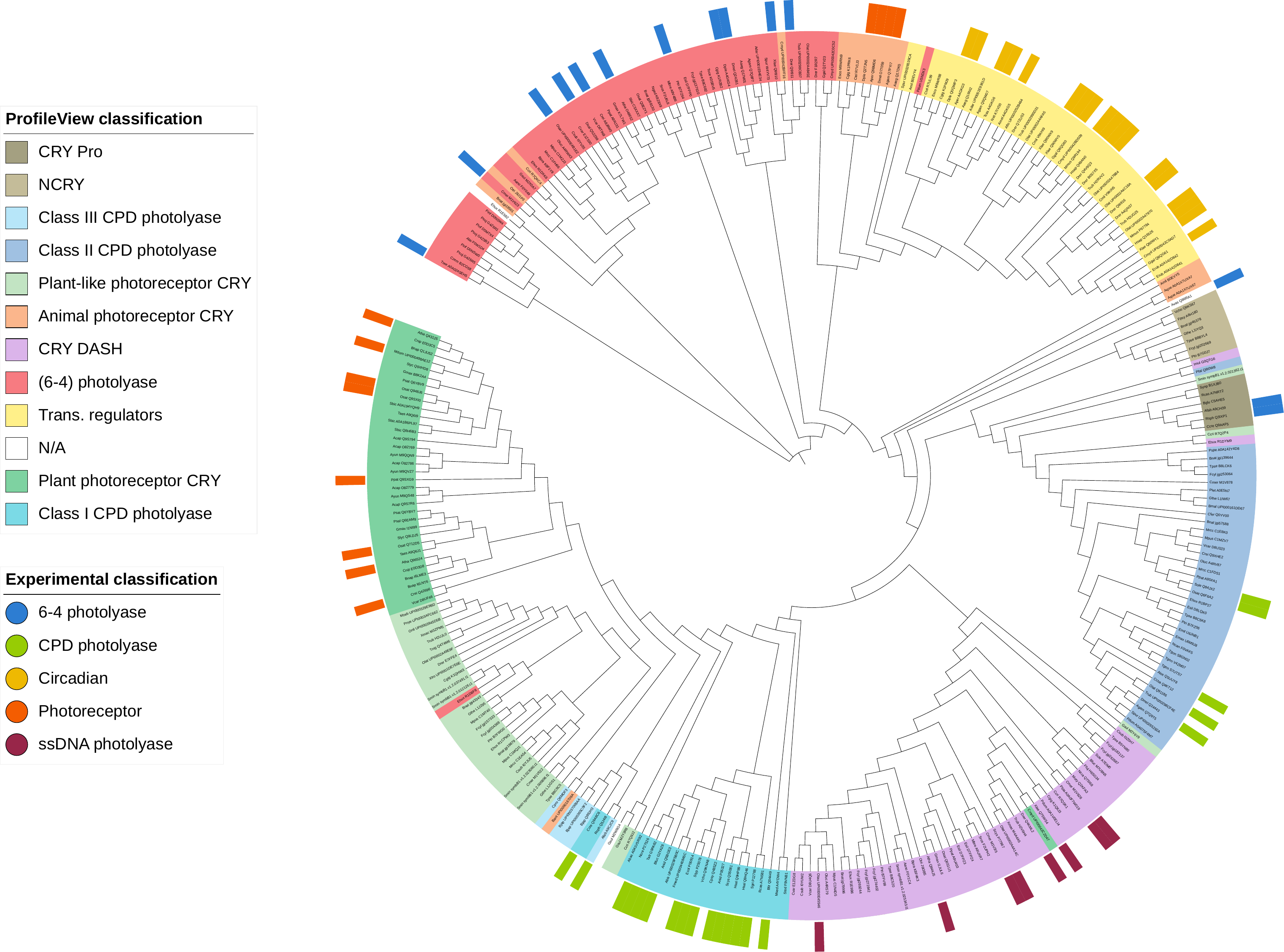}
    \caption{Phylogenetic tree obtained from the CPF protein dataset using the neighbour-joining method and the software \href{http://www.microbesonline.org/fasttree/}{FastTree} with default parameters. The external coloured tags correspond to the experimental classification, whereas the leaf's colour represents the classification inferred by ProfileView in Ref.~\cite{10.1093/molbev/msac070}.}
    \label{fig:FAD phylogenetic tree}
\end{figure*}

\begin{figure*}[ht]
    \centering
    \includegraphics[width=\textwidth]{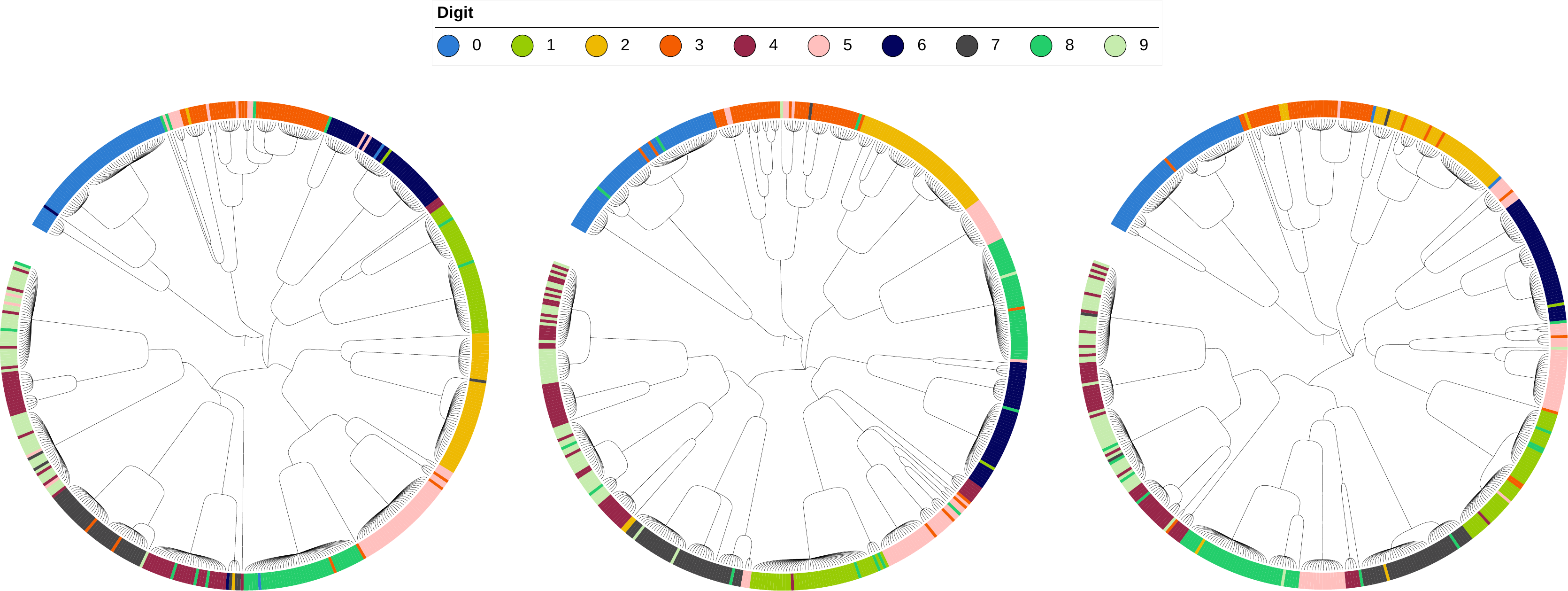}
    \caption{Comparison between three trees of the MNIST dataset generated with three models having the same parameters but trained using different random seeds.}
    \label{fig:trees comparison}
\end{figure*}

\begin{figure*}[ht]
    \centering
    \includegraphics[width=0.8\textwidth]{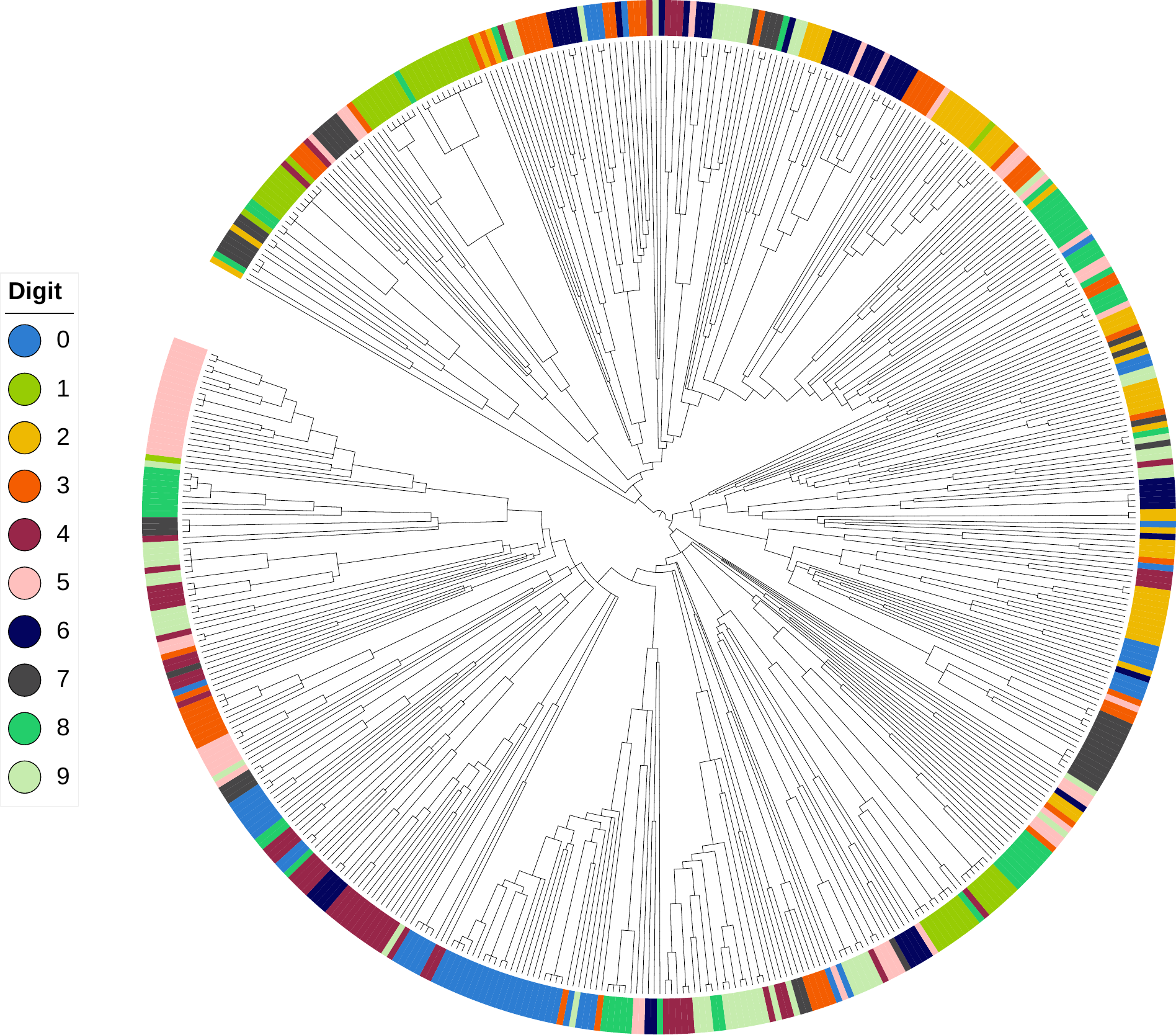}
    \caption{Tree of the MNIST dataset generated through an inverse-annealing of the oldest RBM model. We applied the tree construction procedure to a set of $\sim 100$ models at linearly-spaced inverse temperatures in the interval $\beta \in [0.65, 1]$, starting from the coldest model at $\beta=1$ and progressively increasing the temperature.}
    \label{fig:tree annealing}
\end{figure*}

\begin{figure*}[ht]
    \centering
    \includegraphics[width=\textwidth]{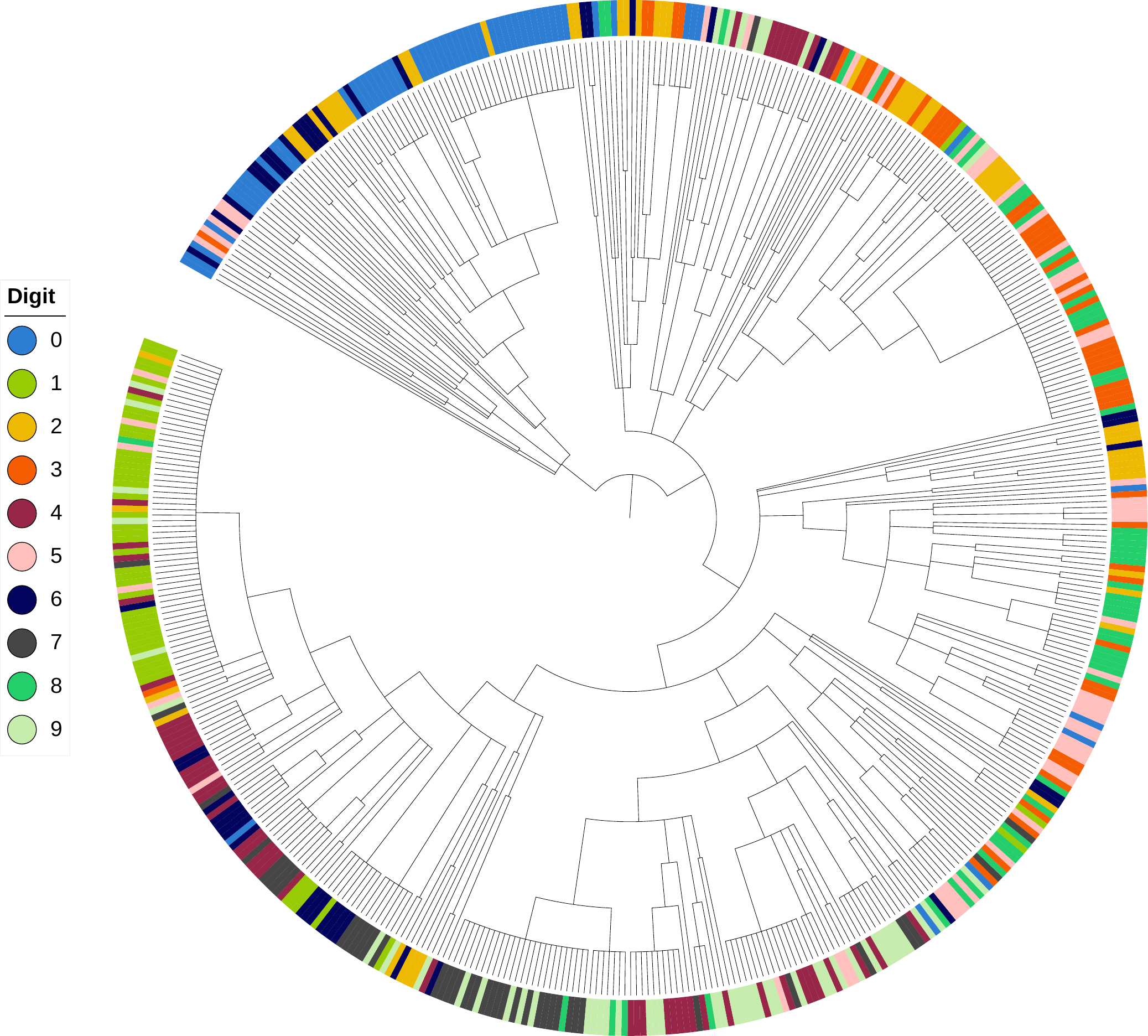}
    \caption{Tree obtained on the MNIST dataset by iterating the TAP equations starting from the data at each age.}
    \label{fig:tree MNIST wrong way}
\end{figure*}

\begin{figure*}[ht]
    \centering
    \includegraphics[width=\textwidth]{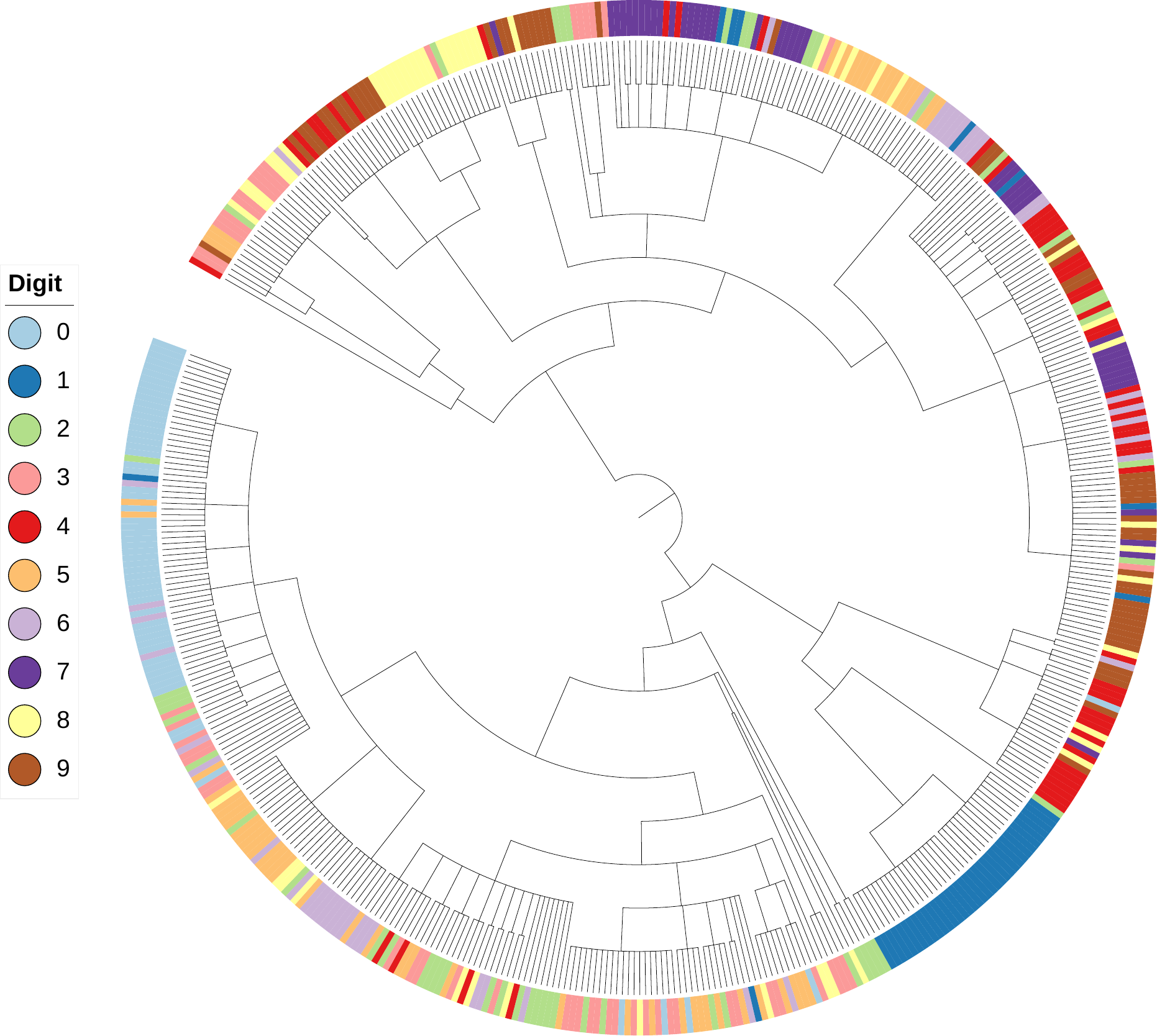}
    \caption{Tree obtained on the MNIST dataset using an RBM trained with PCD-100 using the TAP equations.}
    \label{fig:tree TAP}
\end{figure*}

\end{appendix}



\end{document}